\documentclass[times,twocolumn,final]{elsarticle}
\usepackage{medima}
\usepackage{framed,multirow}

\usepackage{amssymb}
\usepackage{amsmath}
\usepackage{mathrsfs}
\usepackage{latexsym}
\usepackage[linesnumbered, ruled, vlined]{algorithm2e}
\usepackage[normalem]{ulem}
\usepackage{pdfpages}

\usepackage{url}
\usepackage{xcolor}

\usepackage{hyperref}

\definecolor{newcolor}{rgb}{.8,.349,.1}
\usepackage{colortbl}
\usepackage{svg}

\journal{Medical Image Analysis}

\begin{document}

\verso{Manthe \textit{et~al.}}

\begin{frontmatter}

\title{Federated brain tumor segmentation: an extensive benchmark}%
%\title{Federated brain tumor segmentation: a case study\tnoteref{tnote1}}%
%\tnotetext[tnote1]{This is an example for title footnote coding.}

\author[1,2]{Matthis \snm{Manthe}\corref{cor1}}
\cortext[cor1]{Corresponding author:}
\ead{matthis.manthe@insa-lyon.fr}
\author[2]{Stefan \snm{Duffner}}
%\author[1]{Stefan \snm{Duffner}\fnref{fn1}}
%\fntext[fn1]{This is author footnote for second author.}
\author[1]{Carole \snm{Lartizien}}

\address[1]{Univ Lyon, CNRS, INSA Lyon, UCBL, Inserm, CREATIS UMR 5220, U1294, F‐69621 Villeurbanne, France}
\address[2]{Univ Lyon, INSA Lyon, CNRS, UCBL, Centrale Lyon, Univ Lyon 2, LIRIS, UMR5205, F-69621 Villeurbanne, France}

%\received{1 May 2013}
%\finalform{10 May 2013}
%\accepted{13 May 2013}
%\availableonline{15 May 2013}
%\communicated{S. Sarkar}

\begin{abstract}
%%%
%The attention given to federated learning in the medical image analysis field grows year by year. An exponential amount of federated methods to counteract the effect of statistical heterogeneity of decentralized datasets are being published, from both medical and machine learning communities. Different classes of methods emerged, either global (one final model), personalized (one model per institution) or hybrid (one model per cluster of institutions). Their applicability on the recently published Federated Brain Tumor Segmentation 2022 dataset is unknown. We propose an extensive benchmark of federated learning algorithms from all three classes on this task. We show that global methods such as SCAFFOLD, personalized and hybrid methods can all significantly improve models' segmentation performance compared to the baseline FedAvg, while limiting final model(s) bias toward the predominant data distribution of the federation. Moreover, we try to provide a deeper understanding of the behaviour of federated learning on this task through multiple alternative partitioning. Our code is available at (LINK).
Recently, federated learning has raised increasing interest in the medical image analysis field due to its ability to aggregate multi-center data with privacy-preserving properties. A large amount of federated training schemes have been published, which we categorize into global (one final model), personalized (one model per institution) or hybrid (one model per cluster of institutions) methods. However, their applicability on the recently published Federated Brain Tumor Segmentation 2022 dataset has not been explored yet. We propose an extensive benchmark of federated learning algorithms from all three classes on this task. While standard FedAvg already performs very well, we show that some methods from each category can bring a slight performance improvement and potentially limit the final model(s) bias toward the predominant data distribution of the federation. Moreover, we provide a deeper understanding of the behaviour of federated learning on this task through alternative ways of distributing the pooled dataset among institutions, namely an Independent and Identical Distributed (IID) setup, and a limited data setup. Our code is available at (\url{https://github.com/MatthisManthe/Benchmark_FeTS2022}).

%%%%
\end{abstract}

\begin{keyword}
%% MSC codes here, in the form: \MSC code \sep code
%% or \MSC[2008] code \sep code (2000 is the default)
%% Keywords
\KWD Deep learning\sep Federated learning\sep Medical image segmentation\sep BraTS\sep Personalized federated learning\sep Clustered federated learning
\end{keyword}

\end{frontmatter}

%\linenumbers

\section{Introduction}

% Start by federated learning, let glioma for the middle of the intro.
%The segmentation of glioma sub-regions
%S in MRI?
%is of great value for the diagnosis and prognosis of this disease. As it costs a large amount of time for certified radiologists to perform a single segmentation, its automation would greatly improve the pace of the treatment of patients.

Deep learning showed state-of-the-art performance on a variety of medical image analysis tasks \cite{baur_deep_2019, futrega_optimized_2022, maier_isles_2017, zeineldin_multimodal_2022}. However, training precise deep learning models requires a large amount of sensitive and high-quality data. Data centralization possibilities are limited: restrictive legislations such as the General Data Protection Regulation (GDPR) in Europe, the overall sensitivity of medical images, as well as the high labelling cost of each exam by clinical expert, make the aggregation of large amounts of data complicated if not prohibited. 

Federated learning was defined in \cite{mcmahan_communication_efficient_2017} with the seminal algorithm called \textit{Federated Averaging (FedAvg)} as a decentralized privacy-preserving machine learning paradigm enabling numerous institutions to collaboratively train a model on their cumulated data without ever sharing them. Even if the privacy of standard federated learning methods is a field of research in itself \cite{rodriguez_barroso_survey_2023, yang_gradient_2023, zhu_deep_2019}, classical federated learning methods are a significant step towards effective decentralized training conformed with privacy and ethical requirements.

We have seen a growing interest in federated learning in the medical field as it could drastically increase the data quantity used to train a model at a limited privacy cost. It became a popular research topic on a variety of tasks, from fMRI analysis \cite{li_multi_site_2020} to brain anomaly detection \cite{bercea_federated_2022}, based on fully \cite{qu_handling_2022}, weakly \cite{lu_federated_2022, yang_federated_2021}, or un- \cite{wu_distributed_2022} supervised models, even mixing these levels of supervision \cite{wicaksana_fedmix_2022}. There are overall more and more incentives to perform multi-centric research \cite{martensson_reliability_2020, wachinger_detect_2021}, as well as more available public datasets to do so \cite{bilic_liver_2023}. In particular, the Federated brain Tumor Segmentation (FeTS) initiative published the institution-wise partitioning of the BraTS2021 dataset \cite{baid_rsna_asnr_miccai_2021, karargyris_medperf_2021, pati_federated_2021, reina_openfl_2022}, composed of around 1200 multi-modal brain MRI scans of gliomas from 23 clinical institutions along with their multi-label tumor segmentation masks including the whole tumor outline as well as the enhancing and necrotic areas. A significant amount of pioneering works showed that the federated learning paradigm was attractive on this task \cite{sheller_federated_2020, pati_federated_2022}, with two FeTS challenges organized in 2021 and 2022 \cite{pati_federated_2021, isik_polat_evaluation_2022, khan_adaptive_2022, machler_fedcostwavg_2021}, as well as some independent works exploring other aspects of decentralized learning on this valuable segmentation task \cite{camajori_tedeschini_decentralized_2022, li_privacy_preserving_2019}.

%In parallel, an exponential amount of articles are published 
In the past few years, a very large amount of research work has been done by the machine learning community on the general topic of federated learning addressing different limitations. It has been clearly identified now that the statistical distribution of data across institutions, potentially heterogeneous, can drastically alter the performance of standard federated learning methods such as \textit{FedAvg} \cite{li_federated_2020, karimireddy_scaffold_2020}. On this specific problem, a large number of lines of research have been defined, which can be categorized into three classes: global methods \cite{mcmahan_communication_efficient_2017}, i.e. training one model generalizing on the data of every institution; personalized methods \cite{li_ditto_2021}, i.e. training one personalized model for each institution while each benefiting from the federation; and hybrid methods \cite{sattler_clustered_2021}, i.e. building a set of models each designed to perform well on a set of institution or samples. Most of these state-of-the-art (SOTA) methods are outside of the scope of previous works on BraTS dataset and have never been explored on the brain tumor segmentation task yet. Moreover, novel federated methods proposed in both the medical image analysis and machine learning communities mainly focus on patient-level classification tasks, with only few works on segmentation tasks.

We propose a benchmark of several global, personalized and hybrid state-of-the-art federated learning methods on the FeTS2022 dataset. Its purpose is to first investigate the performance of standard federated learning methods on the latest version of the dataset, as well as to highlight which class of methods tend to give better performances considering its data distribution. We complement existing works on FeTS dataset by exploring the application of these methods in a significantly different federated setup, from chosen local optimizer to computational constraints. The main contributions of our work are as follows
\begin{itemize}
    \item We propose, to the best of our knowledge, the first benchmark of personalized and clustered federated learning methods on brain tumor segmentation, showing the potential of both of these paradigms for this task,
    \item We fixed a common estimated time budget enabling the fastest algorithms to reach their validation plateau, analysing results close to convergence.
    \item We analyse the performance biases between institutions brought by each federated optimizer.
    \item We defined each method in a common formal framework with some adaptations for our federated setup, and publicly available implementations.
\end{itemize}
Moreover, we explore additional synthetic partitions of the dataset, namely a randomly redistributed version as well as a version with a limited amount of data per institution with a more controlled heterogeneity. We bring with these secondary experiments a better understanding of the behaviour of federated learning methods on this task.

In section \ref{sec:related_works}, we detail the current trends in both the federated learning community and its application to neuro-image analysis. In section \ref{sec:methods}, we motivate the choice of each method included in this benchmark and describe them. We define the experimental setup in section \ref{sec:exp}, and explore the results in section \ref{sec:results}. We discuss the choices made as well as the limitations of the benchmark and conclude in section \ref{sec:discussion}.

\section{Related work}\label{sec:related_works}

\subsection{Federated learning}
The first published federated learning algorithm was \textit{Federated Averaging (FedAvg)} \cite{mcmahan_communication_efficient_2017}, where each participating institution performs multiple gradient steps locally on a received model between server-side averaging steps. We can underline two main research axes devoted to the improvement of this algorithm in terms of convergence speed and the final accuracy of the obtained global model. 

The first one is to perceive the aggregation server-side as the approximation of a gradient, thus exploring federated algorithms as a branch of standard stochastic optimization theory. Thus was explored, for example, the addition of server-side learning rate, momentum and adaptivity \cite{jhunjhunwala_fedexp_2023, reddi_adaptive_2022}. A second line of work is to perceive the aggregation step as a model fusion task. Some authors proposed to match neurons or channels between local models before averaging the weights by the Hungarian algorithm (\textit{FedMA} \cite{wang_federated_2020}) or graph matching \cite{liu_deep_2022}, or by allocating semantic channel positions prior to training in the network for a better averaging (\textit{Fed$^2$} \cite{yu_fed2_2021}). 

It was quickly shown that the statistical heterogeneity of local datasets could harm the convergence the more local steps are performed between server synchronizations \cite{wang_tackling_2020, zhao_federated_2018}. Numerous authors proposed improvements over \textit{FedAvg} in this specific case. Most of them alter the local optimization to limit the "client drift", namely the quick overfitting of local optimizations towards local optima, negating the efficiency of server aggregations. Proximal regularization was first proposed (\textit{FedProx} \cite{li_federated_2020}), followed by several important methods trying to guide local optimizations away from local optima. Examples are \textit{SCAFFOLD}, leveraging control variates to promote the convergence of local optimizations toward the global optima \cite{karimireddy_scaffold_2020}, or \textit{MOON} based on a local model contrastive loss \cite{li_model_contrastive_2021}. Other authors explored the application of robust optimization methods locally, such as adaptations of Elastic Weight Consolidation (\textit{FedCurv} \cite{shoham_overcoming_2019}) or generative replay (\textit{FedReplay} \cite{qu_handling_2022}) from the continual learning field, or employed local sample synthesis \cite{tang_virtual_2022, wang_federated_2022}. Some authors also used server-side ensemble distillation from auxiliary data to improve the aggregated global model before sending it to participants (\textit{FedDF} \cite{lin_ensemble_2021} or \textit{FedBE} \cite{chen_fedbe_2020}). Finally, some work focuses on neural architectures that are more suited to federated learning, showing that batch normalization is not adapted to local distribution shifts \cite{li_fedbn_2021, wang_why_2023} or that transformer models might be easier to train in a federated setup than convolutional ones \cite{qu_rethinking_2022}.

\subsection{Personalized and hybrid federated learning}
In parallel to the previously described methods, some authors suggested that training a plurality of models instead of training a single global model performing well on the data of every institution could help bridge the gap between centralized and federated performance in the case of high statistical heterogeneity.

The first line of research, named \textbf{personalization}, aims at training one model per institution each performing better than a global model obtained by standard federated learning and than what could be obtained by isolated institutions. As a straightforward example, personalization can be performed by simply fine-tuning locally a model obtained by \textit{FedAvg}, which was shown quite efficient \cite{mansour_three_2020}. The idea of partial model sharing quickly followed, with the earliest methods retaining local the first (\textit{LG-FedAvg} \cite{liang_think_2020}) or the last (\textit{FedPer} \cite{arivazhagan_federated_2019}) local layers, or the batch normalization parameters (\textit{FedBN} \cite{li_fedbn_2020}) of a neural network. Numerous works then followed this line of work (\textit{CD$^2$-FedAvg} \cite{shen_cd2_pfed_2022}, \textit{PartialFixedShare} \cite{pillutla_federated_2022}, \textit{PartialFed} \cite{sun_partialfed_2021}).

The personalization task can be linked to a large number of subfields in machine learning, and multiple formulations were proposed. Authors interpreted personalization in the meta-learning framework, with the objective of training a server-side meta-model easily fine-tunable on local datasets. In that vein, adaptations of the model-agnostic meta-learning algorithm (\textit{MAML} \cite{finn_model_agnostic_2017}) were explored (\textit{PerFedAvg} \cite{fallah_personalized_2020} or \textit{pFedMe} \cite{dinh_personalized_2020}). Alternatively, each local dataset of a federation can be viewed as a task in the multi-task learning framework. Thus, authors tried to adapt multi-task regularization methods to the federated setup, such as \textit{Ditto} \cite{li_ditto_2021} or \textit{FedEM} \cite{marfoq_federated_2022}. 

More specific to the federated setup, some researchers explored the combination of local and global methods to perform personalization. The first works explored the definition of personalization objectives as finding the best interpolation between local and global models (\textit{Mapper} \cite{mansour_three_2020}, \textit{SoftPull} \cite{xu_closing_2022}), while a recent work also proposed to use a combination of a global federated learning model and local kNN models to perform robust personalization (\textit{kNN-Per} \cite{marfoq_personalized_2022}). We can also note an interesting idea of leveraging a hypernetwork server-side to output the parameters (or a subset of them) of personalized models (\textit{pFedHN} \cite{shamsian_personalized_2021}).

Finally, tangentially to multi-task adaptations, some authors explored personalization methods based on the computation of similarities between local datasets through different proxies, either using the loss of a local model on the datasets of other institutions (\textit{FedFOMO} \cite{zhang_personalized_2021}), or distances in the parameter space (\textit{FedAMP} \cite{huang_personalized_2021}) to use them as local aggregation weights. This implicit hypothesis of institutions sharing more similarity with a subset of the federation also motivated a refinement of the personalization framework into a \textbf{clustered} (or hybrid) one. In this framework, the objective is to train a set of models, each specialized for a set of institutions. Cluster federated learning also implies the computation of similarities between institutions: some authors experimented with the loss of the clustered models to incrementally compute clusters of institutions during training (\textit{IFCA} \cite{ghosh_efficient_2020}), or computed clusters based on the cosine similarity matrix between local updates (\textit{CFL} \cite{sattler_clustered_2021}).

\subsection{\label{sec:related_brats} Federated learning in brain tumor segmentation}

Pioneering works already established the potential of federated learning in brain tumor segmentation. A first proof of concept was proposed in \cite{sheller_federated_2020}, where standard \textit{FedAvg} was compared to cyclic weight transfer on BraTS 2018 dataset, highlighting that even in a cross-silo setting, the latter did not scale well due to catastrophic forgetting from one institution to the other, while the former reached a close match in performance with centralized training. This work was rapidly followed by experiments with differential privacy and local momentum restart \cite{li_privacy_preserving_2019}. Furthermore, two editions of the Federated brain Tumor Segmentation Challenge, FeTS 2021 and 2022 were hosted \cite{pati_federated_2021}. They were dedicated to the impact analysis of institution sampling, frequency of server synchronisation and aggregation weights strategies in \textit{FedAvg}. Several novel methods were proposed during this challenge. \cite{isik_polat_evaluation_2022, khan_adaptive_2022, machler_fedcostwavg_2021, machler_fedpidavg_2023} Namely, \textit{FedCostwAvg} \cite{machler_fedcostwavg_2021} was the winner of 2021's edition, weighting the local updates by a linear combination of the local sample size and the local loss improvement from one round to another. The authors further improved their method on the second edition, named \textit{FedPIDAvg} \cite{machler_fedpidavg_2023}, by adding a third term in the weight of the local updates: a form of momentum on the local losses. Other participants experimented with \textit{FedAvg-M}, \textit{FedAdam}, robust aggregation schemes such as median, trimmed mean (discounting local updates that are too far from the median), Top-K mean (discounting local updates that did not sufficiently improve the local loss compared to others), and other forms of loss based server-side aggregations \cite{rawat_robust_2022, tuladhar_federated_2022}. Finally, other authors explored during the challenge the usage of adaptive number of local epochs per round and learning rate decay with \textit{FedAvg} as well as \textit{FedNova} and \textit{FedAvg-M} \cite{isik_polat_evaluation_2022, tuladhar_federated_2022}. Finally, an outstanding real world application of federated learning on 71 institutions worldwide on brain tumor segmentation was recently published \cite{pati_federated_2022}. The authors showed that applying \textit{FedAvg} on such a massive coalition of institutions significantly improves the performance of the final model compared to what could be obtained with only the public BraTS datasets.

In comparison to this first ever challenge in federated medical image segmentation, we made the choice to use a smaller 3D U-Net. Considering our limited computational capacity, we chose the lowest possible training time budget enabling some variants of FedAvg to reach a validation plateau, getting closer to convergence than FeTS challenges. We also focused on recently proposed new classes of SOTA methods such as personalization and clustered methods which were not implementable with the coding constraints of the challenges. We used standard SGD locally instead of Adam to stay consistent with the federated learning literature. In essence, we try to complement the results of the FeTS challenges with a different federated setup. Moreover, along with proposing rigorous experiments of SOTA methods, we hope that our work can help identifying valuable "types" of federated learning methods for large scale real-world implementations such as \cite{pati_federated_2022}.

\section{Methods}\label{sec:methods}

As we could not experiment with every state-of-the-art federated learning methods, we selected the most representative and established ones from different groups of federated techniques (from adaptive and variance reduction-based global methods, finetuning-based or multi-task-based personalized methods, etc.). We define each of them with the potential adaptations which were required to apply them on the 3D brain tumor segmentation task of interest.

\subsection{Notations and centralized problem}

Let $K$ be the number of institutions each with a local dataset $D_k:=\{(x_{k,i}, y_{k,i})\}_{i=1}^{n_k}$, with $x_{k,i}\in\mathcal{X} = \mathbb{R}^{m\times h\times w\times d}$ the multimodal MRI scan to segment, $y_{k,i}\in\mathcal{Y} = \{0,1\}^{l\times h\times w\times d}$ its associated multi-label ground-truth segmentation map, $n_k$ the local dataset size of institution $k$ and $N = \sum_{k=1}^Kn_k$ the total number of samples. We note $w\in\mathcal{W} = \mathbb{R}^p$ the parameters of the neural network to be optimized for the downstream task.

Given a loss function $l: \mathcal{W}\times \mathcal{X}\times \mathcal{Y} \rightarrow \mathbb{R}$, the classical formulation of the neural network optimization problem in a centralized setup can be defined as
\begin{equation}
    w^* = \underset{w\in\mathcal{W}}{argmin } \sum_{i=1}^{N}l(w, x_{i}, y_{i})\label{eq:centralized_obj}
\end{equation} over the pooled dataset $D = \bigcup_{k=1}^{K}D_k$.

\subsection{Global federated learning}
\subsubsection{Problem formulation}
Under privacy constraints, pooling each institution's local dataset together on a single computation unit is prohibited. Federated learning was proposed (\cite{mcmahan_communication_efficient_2017}) to train a neural network over decentralized data in a collaborative fashion such that raw data never leaves data warehouses of each participating institution. The decentralized optimization problem can be defined as 
\begin{equation}
    w^* = \underset{w\in\mathcal{W}}{argmin } \sum_{k=1}^K\sum_{i=1}^{n_k}l(w, x_{k,i}, y_{k,i})\label{eq:global_fed_obj} \; ,
\end{equation}
recovering the centralized one. Explored global federated algorithms are summarized in Algorithm \ref{algo:global}.

\subsubsection{Federated Averaging}\label{sec:fedavg}
\textit{Federated Averaging (FedAvg)} is a pioneer work proposed by \cite{mcmahan_communication_efficient_2017}. They proposed to organize the collaborative training process in several communication rounds. During each round $t$, the aggregation server sends the parameters $w^t$ of the current global model to each participant. They perform $E$ epochs of local training using $w^t$ as an initialization to obtain a local update $\Delta w^t_k = w^t_k - w^t$ (note that in most theoretical works, $U$ gradient steps per institution are used instead). They then send these updates back to the server, which aggregates them in a single global update applied to the global model following
\begin{equation}\label{eq:aggregation_fedavg}
    w^{t+1} = w^t + \sum_{k=1}^Kp_k\Delta w_k^{t} \; ,
\end{equation} with $p_k$ an aggregation weight. Both uniform ($p_k=\frac{1}{K}$) and weighted ($p_k=\frac{n_k}{N}$) averaging were proposed, the latter introduced to ensure convergence towards the global objective of Equation \ref{eq:global_fed_obj}.

As the term \textit{FedAvg} encompasses all of these different variations, we specifically define \textit{FedAvg with fixed local epochs} its variant with each institution performing E local epochs between each communication round and server-side weighted average aggregation, \textit{FedAvg with fixed iterations} its variant with U local updates and weighted averaging and \textit{FedAvg with uniform averaging} its variant with E local epochs and uniform averaging. Fixed batch size is set common to all institutions for all variants.

\subsubsection{FedNova}
In the specific case of an unbalanced amount of samples per institution, the authors of \cite{wang_tackling_2020} underlined an inconsistency in the optimization of \textit{FedAvg} with a fixed number of local epochs per round and weighted aggregation (i.e. $p_k=\frac{n_k}{N}$). With a unique fixed batch size, institutions owning more samples than others will perform more gradient steps, giving implicitly more weight to their local updates in the aggregation. This motivated the development of a framework to limit this problem for a large amount of SOTA methods, leveraging normalized gradient aggregation. In our study, we chose to apply their adaptation of \textit{FedAvg}, following a simplification of the formulation in \cite{isik_polat_evaluation_2022}. FedNova reduces to \textit{FedAvg} with a uniform weighting scheme and an analytical server learning rate $\gamma\geq1$
\begin{equation}
    \gamma = K\sum_{k=1}^Kp_k^2
\end{equation} where the aggregation becomes
\begin{equation}
    w^{t+1} = w^t + \gamma\frac{1}{K}\sum_{k=1}^K\Delta w^t_k
\end{equation}

\subsubsection{SCAFFOLD}
To reduce the variability of local updates and enforce convergence of local optimizations toward the global optima, \cite{karimireddy_scaffold_2020} added the computation of local and global control variates used as a form of momentum during local optimizations. After each local training, an institution updates its local control variate $c_k^t$ following
\begin{equation}
    \Delta c_k^{t} = - c^t + \frac{1}{s_k\eta_l}\Delta w_k^t
\end{equation}
\begin{equation}
    c_k^{t+1} = c_k^t + \Delta c_k^{t}
\end{equation} with $s_k$ the number of gradient steps performed by institution $k$ and $c^t$ a global control variate, and sends them to the server along with the computed local updates $\Delta w_k^t$. In addition to the classical aggregation of FedAvg, the server aggregates control variate updates into the global control variate following
\begin{equation}
    c^{t+1} = c^t + \sum_{k=1}^K p_k\Delta c_k^t
\end{equation} which is sent back to the institutions. A local update step on a batch $b_k$ of samples of institution $k$ is altered by these control variates following Equation \ref{eq:scaffold_update_local} (omitting local step and global round superscripts for readability).
\begin{equation}\label{eq:scaffold_update_local}
    w_k = w_k - \eta_l(\nabla l(w_k, b_k) - c_k + c) .
\end{equation} 
Note that we use the adaptation to batch training of \textit{SCAFFOLD} published in \cite{reddi_adaptive_2022} to deal with imbalanced local dataset sizes.

\subsubsection{FedAdam}
\cite{reddi_adaptive_2022} proposed to extend classical adaptative methods such as Adagrad, Adam and Yogi to federated learning by applying them server-side while giving proofs of their effectiveness in this context. For computational cost reasons, we chose to focus on \textit{FedAdam} only. At aggregation step $t$, the server computes first and second-order momentums
\begin{equation}
    \Delta^{t+1} = \beta_1\Delta^{t} + (1-\beta_1)\sum_{k=1}^K p_k\Delta w_k^t
\end{equation}
\begin{equation}
    v^{t+1} = \beta_2v^{t} + (1-\beta_2)\left(\sum_{k=1}^K p_k\Delta w_k^t\right)^2
\end{equation} and aggregates local updates following
\begin{equation}
    w^{t+1} = w^t + \eta_s\frac{\Delta^t}{\sqrt{v^t}+\tau} \; .
\end{equation}

\subsubsection{q-FedAvg}
Soon after the publication of \cite{mcmahan_communication_efficient_2017}, \cite{li_fair_2020} explored the notion of fairness in federated learning. They showed that \textit{FedAvg} could accentuate the disparities of performance of the global model between institutions. Extending the pioneering work of \cite{mohri_agnostic_2019}, they proposed to give more weight server-side to updates computed by institutions on which the global model at rounds $t$ performs worse than others. Formally, they define the fair federated objective as follows
\begin{equation}
    w^*=\underset{w\in\mathcal{W}}{argmin}\sum_{k=1}^K\frac{p_k}{q+1}\left(\frac{1}{n_k}\sum_{i=1}^{n_k}l(w,x_i^k, y_i^k)\right)^{q+1}
\end{equation} with $q\in\mathbb{R}^+$ a hyperparameter enabling to define the desired level of fairness ($q=0$ is equivalent to \textit{FedAvg}, while $q\rightarrow\infty$ gives agnostic federated learning \cite{mohri_agnostic_2019}, optimizing for the worse performing institution). To optimize this objective, the authors proposed a modification of \textit{FedAvg} called \textit{q-FedAvg}. Each participant weights its local update by the loss of the latest global model on its data to the power $q$ before transmitting it to the server 
\begin{equation}
    \Delta_k^t = \frac{1}{\eta_l}F_k^q(w^t)\Delta w_k^t    
\end{equation} where the local training loss of institution k is defined as follows 
\begin{equation}
    F_k(w):=\sum_{\{x_{k},y_{k}\}\in D_k}l(w,x_{k},y_{k})
\end{equation}
They also compute a regularization value $h_k^t$, used at the server-side in the aggregation step following

\begin{equation}
    h_k^t = qF_k^{q-1}(w^t)||\Delta w_k^t||^2 + \frac{1}{\eta_l}F_k^q(w^t)
\end{equation}
\begin{equation}
    w^{t+1} = w^t + \frac{\sum_{k=1}^K\Delta_k^t}{\sum_{k=1}^Kh_k^t}
\end{equation}

\subsubsection{FedPIDAvg}
The winner of the 2022's edition of the FeTS challenge \cite{pati_federated_2021} proposed \textit{FedPIDAvg} \cite{machler_fedpidavg_2023}, a variation of \textit{FedAvg} with a server-side weighting strategy inspired by PID controllers as follows
\begin{equation}
    w^{t+1} = w^{t} + \sum_{k=1}^K(\alpha p_k + \beta\frac{\Delta l^t_k}{L^t} + \gamma\frac{m^t_k}{M^t})\Delta w^t_k
\end{equation} with $l_{val, k}^t$ the local validation loss of institution $k$ at round $t$, $\Delta l^t_k$ the local validation loss improvement, $m^t_k$ a momentum term on these validation loss values and $L^t$ and $M^t$ normalization terms defined as
\begin{gather}
    l_{val, k}^t = \sum_{i=1}^{n_k^{val}}l(w^{t-1}_k,x_i^k,y_i^k)\\
    \Delta l^t_k =  l_{val, k}^{t-1} - l_{val, k}^t, \ L^t = \sum_{k=1}^K\Delta l^t_k\\
    m^t_k = \sum_{j=0}^5l_{val, k}^{t - j}, \ M^t = \sum_{k=1}^Km^t_k
\end{gather}
Since our federated setup differs from FeTS2022 challenge's one (c.f. Section \ref{sec:related_brats}), we had to slightly modify this method for it to converge properly, only accounting for positive validation loss differences in the weighted average to avoid some gradient "ascent" in the latest stages of training. This resulted in modification of $\Delta l^t_k$ as
\begin{equation}
    \Delta l^t_k =  max(0, l_{val, k}^{t-1} - l_{val, k}^t)
\end{equation}

%Red:FedAdam,
%Green:SCAFFOLD,
%Blue:q-FFL
\begin{algorithm}[h!]
    \SetKwFor{ForEach}{for each}{do}{end}%separate foreach with a space
    \SetArgSty{textrm} %do not automatically use italic in arguments
    \SetKwProg{myalg}{Global Federated Learning \textit{(Server-side)}}{}{}
    
    \myalg{}{
        \DontPrintSemicolon % Some LaTeX compilers require you to use \dontprintsemicolon instead
        \KwIn{$w^0, T, E, \eta_l, $\colorbox[RGB]{250, 200, 200}{$\eta_s, \beta_1, \beta_2, \tau$}, \colorbox[RGB]{200, 200, 250}{$q$}, \colorbox[RGB]{245, 205, 159}{$\alpha, \beta, \gamma$}}

        \colorbox[RGB]{250, 250, 200}{
            \begin{minipage}{0.28\textwidth}
                $\gamma \leftarrow K\sum_{k=1}^Kp_k^2$\;
            \end{minipage}
        }\;
            
        \For{$t \leftarrow 1$ \KwTo $T$}{
            Send $w^t$, \colorbox[RGB]{200, 250, 200}{$c^t$} to each institution\;
            \For{$k \leftarrow 1$ \KwTo $K$}{
                $\Delta w^t_k,  $ \colorbox[RGB]{200, 250, 200}{$\Delta c_k^t$}, \colorbox[RGB]{245, 205, 159}{$l^t_{val,k}$} $\leftarrow Local\_Update(k, w^t, E, $\colorbox[RGB]{200, 250, 200}{$c^t$})\;
            }
            \vspace{1mm}
            \emph{Intermediate computations}\;
            \colorbox[RGB]{250, 200, 200}{
                \begin{minipage}{0.35\textwidth}
                    $\Delta^{t+1} \leftarrow \beta_1\Delta^{t} + (1-\beta_1)\sum_{k=1}^K p_k\Delta w_k^t$\;
                    $v^{t+1} \leftarrow \beta_2v^{t} + (1-\beta_2)(\sum_{k=1}^K p_k\Delta w_k^t)^2$\;
                \end{minipage}
            }\;
            \colorbox[RGB]{200, 250, 200}{
                \begin{minipage}{0.35\textwidth}
                    $c^{t+1} \leftarrow c^t + \sum_{k=1}^K p_k\Delta c_k^t$\;
                \end{minipage}
            }\;
            \colorbox[RGB]{200, 200, 250}{
                \begin{minipage}{0.35\textwidth}
                    $\Delta_k^t \leftarrow \frac{1}{\eta_l}F_k^q(w^t)\Delta w_k^t$\;
                    $h_k^t \leftarrow qF_k^{q-1}(w^t)||\Delta w_k^t||^2 + \frac{1}{\eta_l}F_k^q(w^t)$\;
                \end{minipage}
            }\;
            \colorbox[RGB]{245, 205, 159}{
                \begin{minipage}{0.35\textwidth}
                    $\Delta l^t_k =  max(0, l_{val, k}^{t-1} - l_{val, k}^t)$\;
                    $m^t_k = \sum_{j=0}^5l_{val, k}^{t - j}$\;
                \end{minipage}
            }\;
            \vspace{1mm}
            \emph{Aggregation}\;
            \colorbox[RGB]{250, 200, 200}{
                \begin{minipage}{0.35\textwidth}
                    $w^{t+1} \leftarrow w^t + \eta_s\frac{\Delta^{t+1}}{\sqrt{v^{t+1}}+\tau}$\;
                \end{minipage}
            }\;
            \colorbox[RGB]{200, 200, 250}{
                \begin{minipage}{0.35\textwidth}
                    $w^{t+1} \leftarrow w^t + \frac{\sum_{k=1}^K\Delta_k^t}{\sum_{k=1}^Kh_k^t}$\;
                \end{minipage}
            }\;
            \colorbox[RGB]{200, 250, 200}{
                \begin{minipage}{0.35\textwidth}
                    $w^{t+1} \leftarrow w^t + \sum_{k=1}^{K} p_k\Delta w^t_k$\;
                \end{minipage}
            }\;
            \colorbox[RGB]{250, 250, 200}{
                \begin{minipage}{0.35\textwidth}
                    $w^{t+1} \leftarrow w^t + \frac{\gamma}{K}\sum_{k=1}^{K}\Delta w^t_k$\;
                \end{minipage}
            }\;
            \colorbox[RGB]{245, 205, 159}{
                \begin{minipage}{0.35\textwidth}
                    $w^{t+1} \leftarrow w^t + \sum_{k=1}^{K}(\alpha p_k + \beta\frac{\Delta l^t_k}{\sum_{k=1}^K\Delta l^t_k} + \gamma\frac{m^t_k}{\sum_{k=1}^Km^t_k})\Delta w^t_k$\;
                \end{minipage}
            }\;
            \colorbox[RGB]{200, 200, 200}{
                \begin{minipage}{0.35\textwidth}
                    $w^{t+1} \leftarrow w^t + \sum_{k=1}^{K} p_k\Delta w^t_k$\;
                \end{minipage}
            }
            
        }
        \Return $w^T$\;
    }
    \vspace{2mm}
    \SetKwProg{localupdate}{Local Update \textit{(Institution-side)}}{}{}
    \localupdate{}{
        \DontPrintSemicolon % Some LaTeX compilers require you to use \dontprintsemicolon instead
        \KwIn{$k, w^t, E, $ \colorbox[RGB]{200, 250, 200}{$c^t$}}
        
        $w_k^t \leftarrow w^t$\;
        \For{$e \leftarrow 1$ \KwTo $E$}{
            \For {$b \in \mathcal{B}_k$}{
                $w_k^t \leftarrow w_k^t -\eta_l\nabla l(w_k^t,b)$ \colorbox[RGB]{200, 250, 200}{$+\eta_l(c^t - c_k^t)$}\;
            }
        }
        \colorbox[RGB]{200, 250, 200}{
            \begin{minipage}{0.28\textwidth}
                $\Delta c_k^{t} \leftarrow - c^t + \frac{1}{s_k\eta_l}\Delta w_k^t$\;
                $c_k^{t+1} \leftarrow c_k^t + \Delta c_k^{t}$\;
            \end{minipage}
        }\;

        $\Delta w_k^t \leftarrow w_k^t - w^t$\;
        \colorbox[RGB]{245, 205, 159}{$l^t_{val,k} \leftarrow Local\_Validation(w^t_k)$}\;
        \Return $\Delta w_k^t$, \colorbox[RGB]{200, 250, 200}{$\Delta c_k^t$}, \colorbox[RGB]{245, 205, 159}{$l^t_{val,k}$}
    }
    
\caption{Global federated learning algorithms\\ (\colorbox[RGB]{200, 200, 200}{FedAvg}, \colorbox[RGB]{250, 200, 200}{FedAdam}, \colorbox[RGB]{200, 250, 200}{SCAFFOLD}, \colorbox[RGB]{250, 250, 200}{FedNova}, \colorbox[RGB]{200, 200, 250}{q-FedAvg} and \colorbox[RGB]{245, 205, 159}{FedPIDAvg})}\label{algo:global}
\end{algorithm}

\subsection{Personalized federated learning}
\subsubsection{Problem formulation}
In the context of statistical heterogeneity between institutions' datasets, personalized federated learning emerged as the optimization of one model per institution. The subsequent objective can be formalized as the union of the local objectives

\begin{equation}\label{eq:perso_objective}
    w^*_k = \underset{w_k\in\mathcal{W}}{argmin } \sum_{i=1}^{n_k}l(w_k, x_i^k, y_i^k), \forall k\in\{1, ..., K\}
\end{equation} while each model $w_k$ still benefits from the whole federation. 

\subsubsection{Local training}
The simplest approach fitting the latter objective is for each institution to train a model locally without any communication. 

\subsubsection{Local finetuning}
One can also perform personalization by first training a global model using a global solution such as \textit{FedAvg}, which is then finetuned locally by each institution for several epochs.

\subsubsection{Ditto}
In \cite{li_ditto_2021}, the authors inspired themselves by the field of multi-task learning, where the local dataset of each institution can be seen as a task. They define their regularized bi-level objective for institution $k$ as follows
\begin{equation}\label{eq:ditto_objective}
    w_k^* = \underset{w_k\in\mathcal{W}}{argmin\ } \left(\sum_{i=1}^{n_k}l(w_k, x_i^k, y_i^k) + \frac{\lambda}{2}||w_k - w^*||^2\right)
\end{equation} such that $w^*$ solves the global problem of Equation \ref{eq:global_fed_obj}. They argue that this problem can be solved either by local finetuning over the regularized local objective of Equation \ref{eq:ditto_objective}, or by adding a personalization module in parallel to global training. Since they showed that the latter only provides benefits in case of the presence of malicious users while drastically increasing local computation loads, we implement \textit{Ditto} as the former.

\subsubsection{Partial model sharing}
A large amount of research work has been devoted to the exploration of partial model sharing in federated learning in the literature (Section \ref{sec:related_works}). We chose to focus in this work to the most established ones, namely \textit{FedPer} \cite{arivazhagan_federated_2019} and \textit{LG-FedAvg} \cite{liang_think_2020}. Both share the same principle: the model parameters $w$ are partitioned into federated ones $w_g$ and local ones $w_l$ such that $w = w_g \cup w_l$. Based upon \textit{FedAvg}, local updates are computed similarly and only the federated weights ${w_g}_k$ of institution k are transmitted to the server, aggregated and sent back to the institutions while the weights ${w_l}_k$ are kept completely local and preserved from one local optimization to the other. In \textit{FedPer} the authors proposed to keep the last layers of a model private, while in \textit{LG-FedAvg} the first layers are kept local.

\subsection{Clustered federated learning} 
\subsubsection{Problem formulation}
Depending on the task, assumptions on the existence of clusters of institutions with similar data distributions can be made. Consequently, some authors \cite{ghosh_efficient_2020, sattler_clustered_2021} proposed to define a clustered federated learning objective as the union of each cluster objective
\begin{equation}
    w^*_c = \underset{w_c\in\mathcal{W}}{argmin } \sum_{k\in c}\sum_{i=1}^{n_k}l(w_{c}, x_i^k, y_i^k), \forall c\in \mathscr{C}
\end{equation} with $\mathscr{C}\subset\mathcal{P}(\{1, ..., K\})$ a partitioning of the $K$ institutions, and $w_c$ the model parameters of cluster $c$, while each cluster $c$ benefits from the whole federation.

\subsubsection{Prior clustering}
If possible, one can leverage prior knowledge on the distribution of the data between institutions, effectively defining clustering before federated training. This leads to $\mathscr{C}$ being predefined before federated training while implying the leakage of potentially sensitive information. In the case of our study, we do know that gliomas can be classified as of low or high grade (LGG or HGG respectively, c.f. Section \ref{sec:dataset}). We name \textit{Prior CFL HGG/LGG} the federated finetuning (using \textit{FedAvg}) of the best obtained \textit{FedAvg} model on two clusters of institutions: the ones only owning volumes displaying a low-grade tumor (12, 13, 14 and 15) from the others.

\subsubsection{Sattler's Clustered federated learning (CFL) \cite{sattler_clustered_2021}}
If no prior knowledge can be shared independently to federated learning, cluster assignments can be performed in a variety of ways during training. In \cite{sattler_clustered_2021}, the authors proposed to recursively split clusters of institutions into two during federated training. The method starts with standard FedAvg on a single cluster $c_0 = \{1, ..., K\}$ containing every institution. A list of clusters of institutions $\mathscr{C}$ is saved during the whole optimization process, initialized with the initial cluster $c_0$. At an arbitrary communication round $t$, FedAvg is performed independently in parallel on each cluster of the list. At each communication round, the server assesses if each cluster $c$ in the list $\mathscr{C}$ must be split into two parts $c_1$ and $c_2$ before the next round through the following conditions
\begin{align}
    ||\frac{1}{|c|}\sum_{i\in c}\Delta w_i^t|| < \epsilon_1\\
    \underset{i\in c}{max}\ ||\Delta w_i^t|| > \epsilon_2
\end{align} If so, the server splits the cluster $c$ of institutions in two based on the cosine similarity matrix of received local updates from these institutions
\begin{equation}
    a_{i,j}^t = \frac{\langle \Delta w_i^t, \Delta w_j^t\rangle}{||\Delta w_i^t||||\Delta w_j^t||}, i,j\in c
\end{equation} effectively solving 
\begin{equation}
    c_1, c_2 = \underset{c_1\cup c_2=c}{argmin }\ (\underset{i\in c_1, j\in c_2}{max}a_{i,j})
\end{equation} and splitting cluster $c$ into $c_1$ and $c_2$ if  
\begin{equation}
    \gamma_{max} < \sqrt{\frac{1-\underset{i\in c_1, j\in c_2}{max} a_{i,j}}{2}}
\end{equation}
As the hyperparameters $\epsilon_1, \epsilon_2$ and $\gamma_{max}$ defined to choose which cluster to split at which global round are very hard to tune, we slightly simplify their method. Instead of using $\epsilon_1, \epsilon_2$ and $\gamma_{max}$, we only define as a hyperparameter the communication rounds at which each cluster must be split in two, which enables us to avoid relying on local updates norms which can be very heterogeneous in our extremely unbalanced setup.

\section{Experiments}\label{sec:exp}

\subsection{Federated brain tumor segmentation 2022 challenge dataset}\label{sec:dataset}
Experiments were led on \textit{the MICCAI's Federated Brain Tumor Segmentation 2022 Challenge dataset (FeTS2022)} (\cite{bakas_advancing_2017, pati_federated_2021, reina_openfl_2022}). This dataset is based on the Brain Tumor Segmentation 2021 Challenge dataset (BraTS2021). It consists of 1251 multi-modal brain MRI scans (T1, T1ce, T2 and FLAIR) of size $240\times240\times155$ with isotropic 1mm$^3$ voxel size along with their multi-label tumor segmentation masks including 4 labels, namely background, enhancing tumor (ET), tumor core (TC) and whole tumor (WT). 

\begin{figure}[h!]
\centering
\includegraphics[scale=.255]{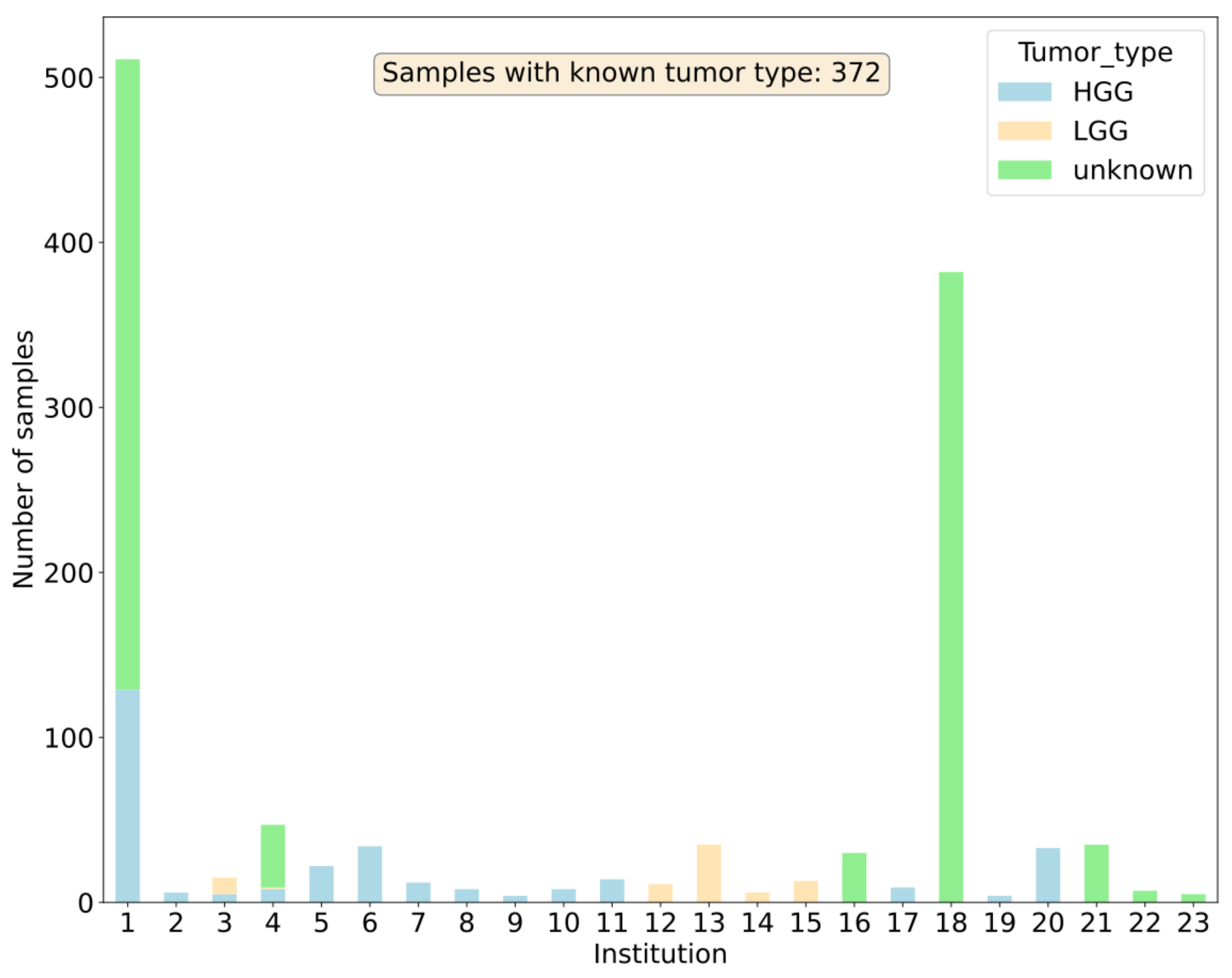}\caption{Number of samples and tumor grade distribution per institution in the challenge setup of FeTS2022 dataset.}\label{fig:tumor_type_skew_fets2022}
\end{figure}

\begin{figure*}[h!]
\centering
\includegraphics[scale=.47]{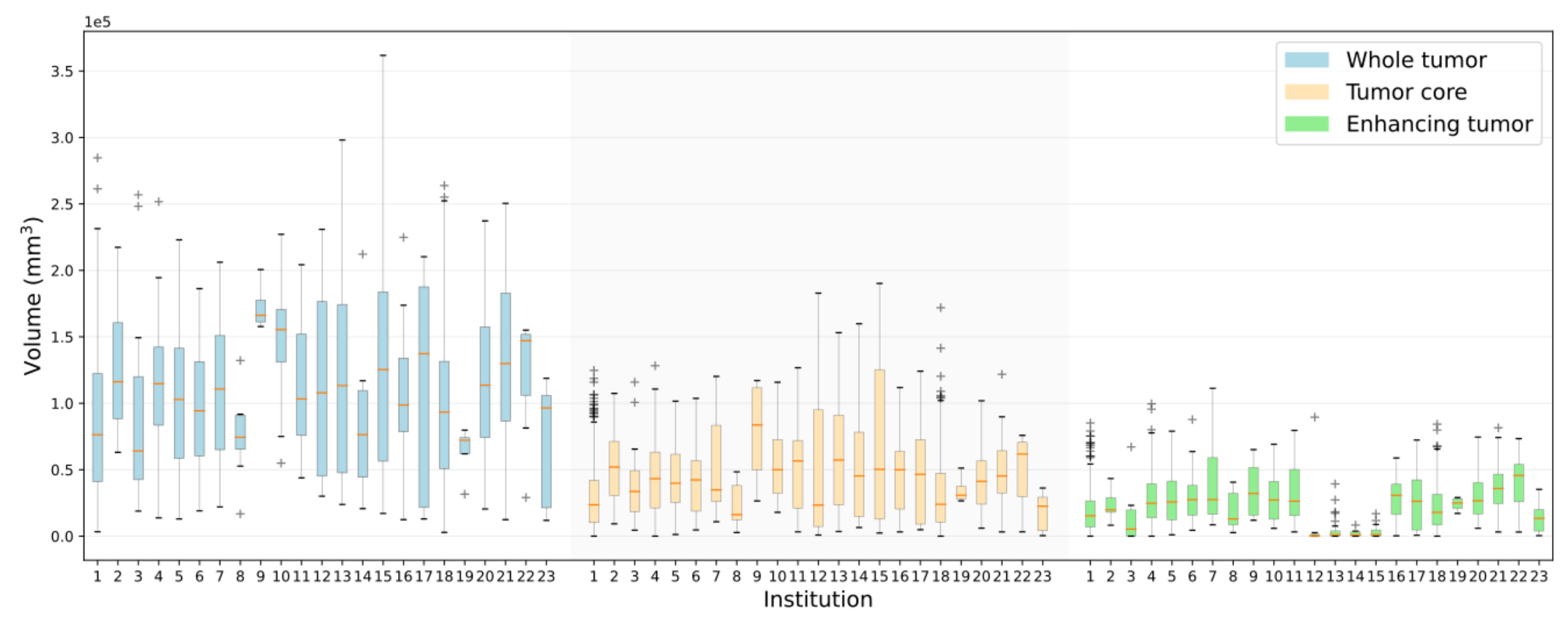}
\caption{Label's volume distribution per institution in FeTS2022 dataset.}\label{fig:label_skew_fets2022}
\end{figure*}

The real-world partitioning along the 23 acquiring institutions is provided in addition to the samples, enabling the simulation of federated learning. Tumors present on each volume can be of low or high grade (LGG or HGG respectively), defining its level of aggressiveness. We can highlight significant heterogeneity between samples of different institutions in terms of quantity, label distribution and tumor grade. Figure \ref{fig:tumor_type_skew_fets2022} shows the distribution of tumor types for each institution. Institutions 1 and 18 provide a large majority of the samples, accounting for 71\% of the total dataset alone while more than 61\% of the institutions own less than 15 samples each. This gives an extremely unbalanced setup. Also note that the tumor grade is known for only $\sim30\%$ of the patients. We know however that institutions 12, 13, 14 and 15 only own LGG samples, institutions 3 and 4 own samples of both grades, and most other institutions only own HGG samples. Figure \ref{fig:label_skew_fets2022} shows the distribution of volumes of each label for each institution. The label distribution is not completely homogeneous across institutions, but it is a priori hard to assess if the label shift we see is sufficient to disturb federated training. We can, however, note that samples of institutions 12, 13, 14 and 15 tend to have smaller ET labels, confirmed by the remark on LGG in \cite{bakas_identifying_2019} that "they do not exhibit much contrast enhancement, or edema". Finally, it must be noted that we do not have access to information about the acquisition protocols of each institution (scanners, parameters, etc.), a feature shift also exists in the dataset.

\subsection{Exploring FL methods on FeTS2022 original partitioning}

In this first group of experiments, we explore the performance of different training schemes on the full dataset in the challenge setup described in Section \ref{sec:dataset}.

\subsubsection{Performance of centralized and institution 1 training} As non-federated baselines, we trained a model with standard SGD on the pooled dataset, which we call \textit{Centralized} training. Since Institution 1 owns more than 500 volumes, it is by itself capable of training a decent model. We thus train a model on its samples alone, which we name \textit{local institution 1} training, to assess in what manner centralized training and federated methods improve this independent model.

\subsubsection{Performance of FedAvg variations} 
As detailed in section \ref{sec:fedavg}, there remains ambiguity about the amount of computation each institution must perform between each communication round in the literature. While theoretical papers such as \cite{karimireddy_scaffold_2020, li_convergence_2020, wang_unified_2022} mostly assume a fixed number of gradient steps per institution, in practice most developed methods such as \cite{mcmahan_communication_efficient_2017, reddi_adaptive_2022} implement a fixed number of epochs per institution as described in Algorithm \ref{algo:global}. In a context of unbalanced data quantity across institutions and fixed batch size, the notion of an epoch is ill-defined, with institutions owning more samples performing more gradient steps than others, potentially giving larger updates. Moreover, in their pioneering work, \cite{mcmahan_communication_efficient_2017} introduces both uniform and weighted server-side aggregations, the latter being introduced to ensure convergence toward the correct optima of Equation \ref{eq:global_fed_obj}. We first explore these ideas experimentally to highlight the best way of distributing computation across institutions with standard \textit{FedAvg} in the challenge setup.

\subsubsection{Performance of global, personalized and hybrid solutions} \label{sec:sota_fets_challenge} We then explore the performance of SOTA federated learning methods described in section \ref{sec:methods} compared to the baselines of \textit{local training}, \textit{centralized training} and \textit{FedAvg}.

\subsection{Performance on alternative data partitionings}
Since label and feature shifts and the unbalanced data quantity distribution of the FeTS2022 dataset can alter the behaviour of federated learning algorithms, we complement the proposed benchmark with the application of federated learning methods on two different synthetic partitionings. 

\subsubsection{IID setup}\label{sec:iid_case}
To understand the effect of the heterogeneous data distribution of the original partitioning, we explore the performance of \textit{FedAvg} while redistributing every sample from every institution randomly in an i.i.d and balanced fashion to 23, 10 and 5 synthetic institutions. We call this \textit{the IID setup}.

\subsubsection{Limited data setup}\label{sec:limited_case}
\begin{figure}[h!]
\centering
\includegraphics[scale=.26]{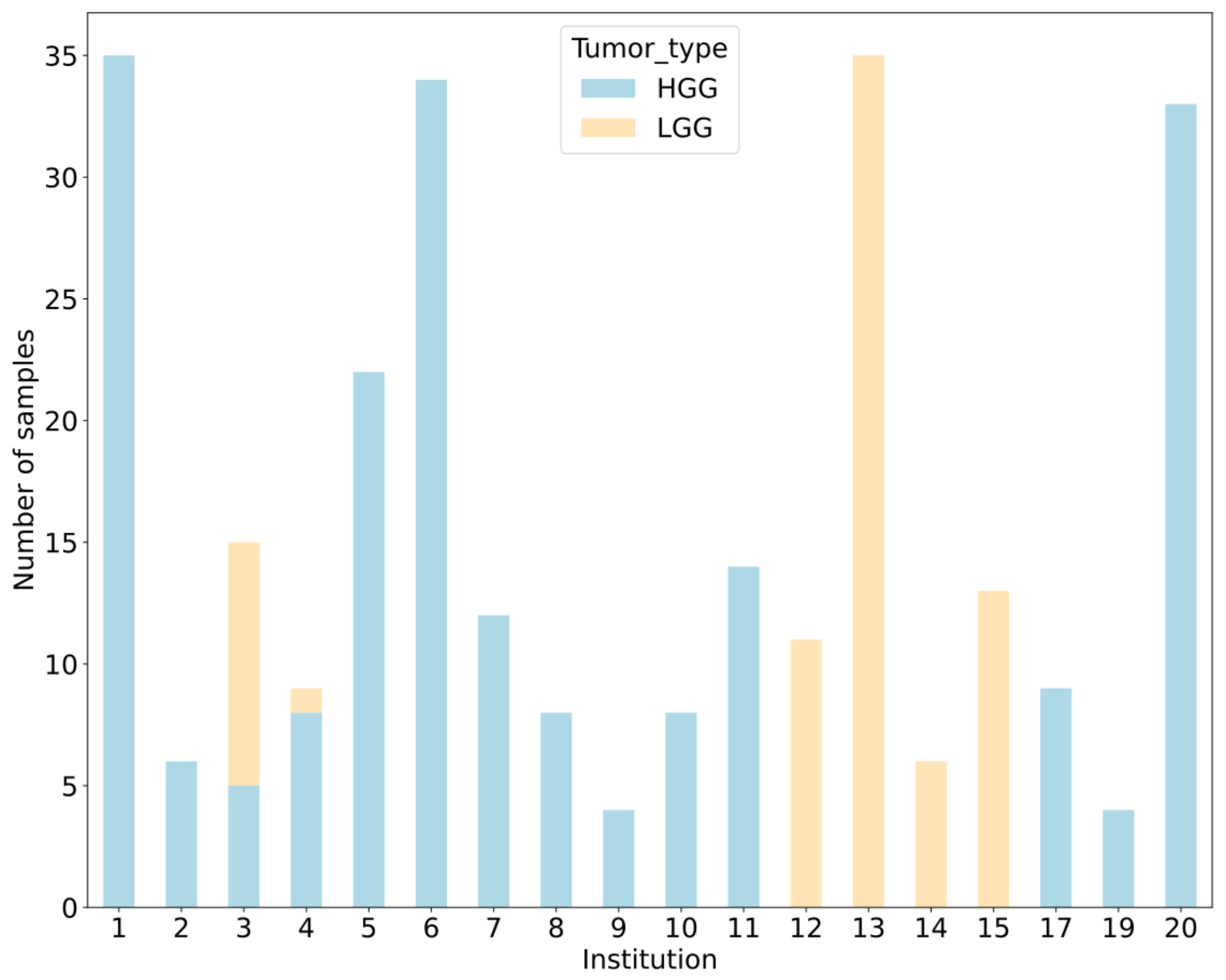}
\caption{Number of samples and tumor grade distribution per institution in the \textit{limited setup} of FeTS2022 dataset. The complete federated subset is composed of 278 patients.}\label{fig:tumor_type_skew_known_tumor}
\end{figure}

It took close to 10 years for the organizers and data curators of BraTS datasets to accumulate such an amount of quality data. While this specific case is of great interest, one can argue that the data distribution, with institutions 1 and 18 owning a very large amount of samples compared to the others, might not be representative of most federated learning applications in a medical cross-silo environment. Moreover, we do not have information on the tumor grade for most samples added to the last version of the dataset, as shown in Figure \ref{fig:tumor_type_skew_fets2022}, which limits our ability to study the potential impact of label shift. For these reasons, we propose first to remove every sample for which we do not have the latter information. This effectively removes institutions 16, 18, 21, 22 and 23, reducing the number of institutions to 18. While doing so, institution 1 would still own a large number of samples compared to others. We thus randomly remove samples from this institution such that it owns only 35 samples, matching the number of samples of the three other main institutions. The final data distribution in this setup is presented in Figure \ref{fig:tumor_type_skew_known_tumor}, named the \textit{limited setup}. LGG tumors are now less marginal representing around 30\% of the dataset, and data is spread in a more balanced fashion across institutions with overall a limited amount of samples per institution. As experimenting with every implemented method would be too expensive, we chose to limit performance analysis on this setup to classical baselines as well as the best-performing methods in section \ref{sec:sota_fets_challenge}, namely the global SCAFFOLD method and the pluralist HGG/LGG CFL method.

\subsection{Experimental setup}

\subsubsection{Model and loss function}
As the backbone model, we chose a classical 3D-UNet \cite{ronneberger_u_net_2015} using Monai's DynUnet implementation (\cite{cardoso_monai_2022, futrega_optimized_2021}) with kernel sizes $3\times3\times3$, and 4 downsampling paths starting from 32 channels after two input convolutional layers up to 512 channels in the bottleneck of the network, doubling the number of channels after each downsampling layer. The network is composed of around 22.5M parameters. Following the architecture choices in previous versions of BraTS challenge \cite{futrega_optimized_2021} and nnUNet \cite{isensee_nnu_net_2021}, we use instance normalization layers without scaling parameters. This is even more motivated by the fact that batch normalization can be intrinsically problematic in a federated setup both from performance \cite{li_fedbn_2021} and privacy \cite{hatamizadeh_gradient_2023} points of view. We use LeakyReLU activations with a negative slope of 0.01 in the whole network. The optimized loss function is a 3D soft Dice loss defined as follows
\begin{equation}
    1-\frac{2*\sum_i^Pp_ig_i + \epsilon}{\sum_i^Pp_i+\sum_i^Pg_i+\epsilon}
\end{equation} where $p_i\in[0,1]$ and $g_i\in\{0, 1\}$ are respectively the prediction and ground truth for voxel $i$, and $\epsilon = 1$ is a smoothing parameter enabling to deal with empty ground truths for TC and ET in some scans. Optimization is performed using patches of size $128^3$ on the task labels (ET, TC and WT) directly. We say that training has been performed for one epoch when one patch per patient has been processed. Inference on a whole multi-modal volume is performed by sliding window inference method with an overlap of 0.5 between patches in gaussian mode.

\subsubsection{Data pre and post-processing}
To limit the computational cost of training and inference, we performed, preliminary to any patching step, a per-volume foreground crop based on the non-zero image masks, followed by a zero-padding to a minimum size of 128 on each dimension. Then, to deal with the heterogeneity of image textures induced by acquisitions on different vendor machines or with different acquisition parameters from the different institutions, each modality of each volume was independently normalized with z-score standardization. Whenever a patient is selected in a training batch, a patch of size $128^3$ is randomly extracted from its associated cropped volume.\\
We avoided using post-processing steps of the output segmentation maps as in nnUNet \cite{isensee_nnu_net_2021}. They would hide imperfections of segmentations from one training algorithm to another, potentially invalidating our conclusions.

\subsubsection{Data augmentation}

We further apply a variety of data augmentations on the fly during training. Precisely, the following augmentations are applied on an extracted patch, in order: 
\begin{itemize}
    \item \textit{Random flip} on each axis with probability 0.5,
    \item \textit{Additive Gaussian noise} with a standard deviation of 0.25 with probability 0.15,
    \item \textit{Gaussian smooth} with a random standard deviation sampled in $[0.5, 1]$ for each dimension with probability 0.15,
    \item \textit{Intensity scaling} with factor randomly sampled in $[0.7, 1.3]$ with probability 0.15,
    \item \textit{Random contrast adjustment} with gamma randomly sampled in $[0.5, 2]$ with probability 0.3.
\end{itemize}

Note that we do not use spatial augmentations such as random zooms and rotations as in nnUNet \cite{isensee_nnu_net_2021}. Preliminary experiments tended to show no benefit in performance when using them, while significantly slowing down training as they are relatively expensive augmentations requiring interpolation.

\subsubsection{Federated cross-validation}
We adapted the paradigm of 5-fold cross-validation to the federated setup to strengthen the statistical significance of the experimental results. This is essential as most institutions do not hold more than 40 volumes. To preserve the representativity of each institution in each decentralized fold, we chose to compute 5 folds on the local dataset of each institution such that a complete decentralized fold is composed of the union of one local fold per institution. Each complete fold serves once as a testing fold. The remaining four other folds are partitioned into a training and validation set following the same principle of locally splitting data into training and validation sets with a proportion of 80-20\% for each institution such that the complete decentralized training and validation sets are the union of each local corresponding set. We use the validation sets for hyperparameter optimization and best model selection during training, such that the testing folds effectively give a glimpse of the generalization performances.

\subsubsection{Evaluation metrics}
Following the choice made by BraTS challenges' organizers, we use a combination of Dice score and Hausdorff distance to evaluate our models. Considering a prediction $P\in\{0,1\}^{l\times h \times w\times d}$ and the ground truth $G\in\{0,1\}^{l\times h \times w\times d}$, the Dice score measures the overlap between them

\begin{equation}
    DICE(P,G) = \frac{2*|P\cap G|}{|P| + |G|}
\end{equation} while the 95\% Hausdorff distance serves as a shape similarity metric, penalizing a significant amount of outlier voxels in the prediction

\begin{equation}
    95HD(P, G) = \mathcal{P}_{95}\Bigl(\{||p, G||_2\}_{p\in P}\cup\{||g, P||_2\}_{g\in G}\Bigl)
\end{equation} with $\mathcal{P}_{95}$ symbolizing the 95th-centile. In the case that a label is not present in a volume, occuring for the ET and TC labels of some samples, the Dice score on this label becomes binary while the Hausdorff distance is undefined. In this case, we only include the Dice score in the average metrics, not the Hausdorff distance. For each metric, we record the average and standard deviation over all test cases. Aggregating institution-averaged metrics would give a similar weight in the evaluation to any two institutions which own a different number of samples, giving a fairness-oriented evaluation in the sense of \cite{li_fair_2020}. We chose to aggregate metrics on a per-sample basis corresponding to the classical centralized evaluation and the actual federated optimization objective, while exploring the fairness aspects by also analyzing performances per institution.

\subsubsection{Evaluation on BraTS2021 Validation dataset}
To ensure the competitiveness of the chosen network architecture and training scheme, we also validate the models trained in a centralized fashion on BraTS2021 validation dataset through the continuous evaluation portal provided by the organizers of the challenges. The ensemble of the five centralized models, each trained on a different set of folds, by a classical logit average is evaluated this way.

\subsubsection{Benchmark budget and fairness}
Guaranteeing the fairness of the benchmark is non-trivial as we allow for different computational and communication loads between methods without necessarily leading every algorithm to convergence due to our computational limitations. We recorded, for each method, their communication cost as well as the amount of total and parallel SGD steps performed during training respectively defined as
\begin{gather}
    S_{total\_SGD\_steps} = R\sum_{k=1}^Ks_k\\
    S_{total\_parallel\_SGD\_steps} = R * \underset{k\in\{1, ..., K\}}{\text{max}}s_k 
\end{gather} with $s_k$ the amount of steps performed by institution $k$ during one round and $R$ the total number of rounds. We also computed a rough estimate of ideal simulated federated training time, inspired by the time simulation of FeTS challenges. We considered the total time taken by a method to be 
\begin{equation}
    T_{total} = R*\text{max}_{k\in[K]}(T^{computation}_k + T^{communication}_k)
\end{equation} The computation time of an institution $k$ for a round is composed of its training steps and its validation of the global model
\begin{equation}
    T^{computation}_k = s_k*t_k^{batch} + n_k^{eval}*t_k^{eval}
\end{equation} with $n_k^{eval}$ the number of validation patients, and $t_k^{batch}$ and $t_k^{eval}$ the training time for one batch and validation time for one patient. Its communication time is composed of its download and upload time
\begin{equation}
    T^{communication}_k = S_{model}*t_k^{download} + S_{model}*t_k^{upload}
\end{equation} with $S_{model}$ the size of the model, and $t_k^{download}$ and $t_k^{upload}$ the net download and upload speeds.

We estimated $t_k^{batch}$ and $t_k^{eval}$ to be 1.86s/batch and 0.80s/patient through multiple centralized epochs using a single V100 GPU giving a realistic computation time of an institution with an extremely high quality system, and $t_k^{download}$ and $t_k^{upload}$ to be 20MB/s and 13.3MB/s based on the fastest communicating institution in FeTS challenge\footnote{Communication time measures were stored in \href{https://github.com/FeTS-AI/Challenge/blob/main/Task_1/fets_challenge/experiment.py}{the source code of FeTS2022 challenge} by the organizers.}. The time simulation and these estimated values are only used as a reference for idealistic conditions of federated learning for a relative fairness of the benchmark, and are discussed in the Discussion section \ref{sec:discussion}. With these assumptions, we trained the models for around 20h of simulated time with each algorithm. It corresponds to approximately 300 rounds of FedAvg, which was enough for its fastest variant to be close to convergence. The hyperparameters of all other algorithms were set accordingly as described in the next section.

\subsubsection{Hyperparameters}
\textit{Centralized} and \textit{Institution 1 local} models were trained for 300 epochs. This number was shown sufficient for the centralized model to be close to convergence i.e. at
%It was enough  
the beginning of the validation plateau from where it would require an order of magnitude more epochs to potentially improve performance in a slightly significant way. For every federated method, we fixed the number of local epochs to 1. For \textit{FedAvg w/ fixed local iterations}, we fixed the number of local iterations to 10 and trained for 720 communication rounds. Other global solutions as well as \textit{FedPer, LG-FedAvg} and \textit{CFL} were trained for 300 communication rounds.  It was enough for the fastest converging methods to reach its validation plateau. \textit{Local finetuning} and \textit{Ditto} were trained locally for 30 epochs, starting from the best \textit{FedAvg} model. In \textit{CFL}, the federation was split in two at communication round 200. Both clusters of \textit{HGG/LGG CFL} were trained for 30 communication rounds starting from the best \textit{FedAvg} model after 270 rounds.

Optimizations were performed using classical SGD with a batch size of 4 and a weight decay of $10^{-5}$ (the latter was shown to be beneficial in \cite{wang_federated_2020} and \cite{yuan_2022_what}). Hyperparameter optimization was performed by grid search using the train-validation split of the first four folds only (not on every fold), apart from \textit{Local finetuning} and \textit{Ditto}, to limit the computational cost of the benchmark. \textit{q-FedAvg} is the only method for which the  hyperparameters selection was manual and not based on the best validation performance. As a fairness-oriented method, each value of $q>0$ decreased the overall performance, we arbitrarily chose one to illustrate its effect. Apart from \textit{q-FedAvg}, an exponential learning rate decay was used multiplying the local learning rate (not the server one if applicable) by 0.995 after each epoch or communication round, roughly dividing by 5 the initial learning rate after 300 epochs. Search spaces and selected values of each hyperparameter for each method with each federated partitioning can be found in Table \ref{tab:hyperparameters_fets2022} of \ref{app:hyperparam}.

\subsubsection{Implementation}
All methods were implemented in Python using Pytorch \cite{paszke_pytorch_2019} and Monai \cite{cardoso_monai_2022}. Training and testing were simulated using 2 NVIDIA V100 32Go GPUs. Training a centralized model for 300 epochs took around 30 hours in this multi-GPU setup.

\section{Results}\label{sec:results}
As results for a large number of methods are presented throughout this section, all-vs-all hypothesis testing would require too much correction for it to be beneficial. We instead chose to perform a non-parametric one-tailed Wilcoxon signed-rank test to support each conclusion taken from the results. We provide in parenthesis the p-value associated with each test.

\subsection{Centralized performance on the BraTS2021 validation data}
We first provide the results of the evaluation of the ensemble of five centralized models through the continuous evaluation portal of BraTS2021. This ensemble of models obtained average Dice scores of 0.920, 0.861 and 0.831 and 95\% Hausdorff distances (in mm) of 4.744, 9.835 and 18.048 for Whole Tumor, Tumor Core and Enhancing Tumor labels respectively. While each model was trained on around 15\% fewer samples than a classically trained model in the challenge setup due to the additional validation sets isolated from training sets, evaluated performances remain close to what could be obtained by the ensemble of larger nnU-Nets which was part of the 2022 challenge's test winner submission  \cite{zeineldin_multimodal_2022} and achieved 0.9213, 0.8718 and 0.8402 Dice scores and 3.82, 8.95 and 16.03 95\% Hausdorff distances. We thus emphasize that the proposed benchmark is not performed using a small model, and most exposed results should be translatable to the practical implementation of federated learning on this task.

\subsection{Exploring FL methods on FeTS2022 original partitioning}

We focus in this section on the performance achieved by the selected SOTA federated methods presented in Section \ref{sec:methods} on the original data partitioning (23 institutions) of FeTS2022.

\subsubsection{Performance of baselines and FedAvg variations}

\begin{figure*}[!t]
\centering
\includegraphics[scale=.235]{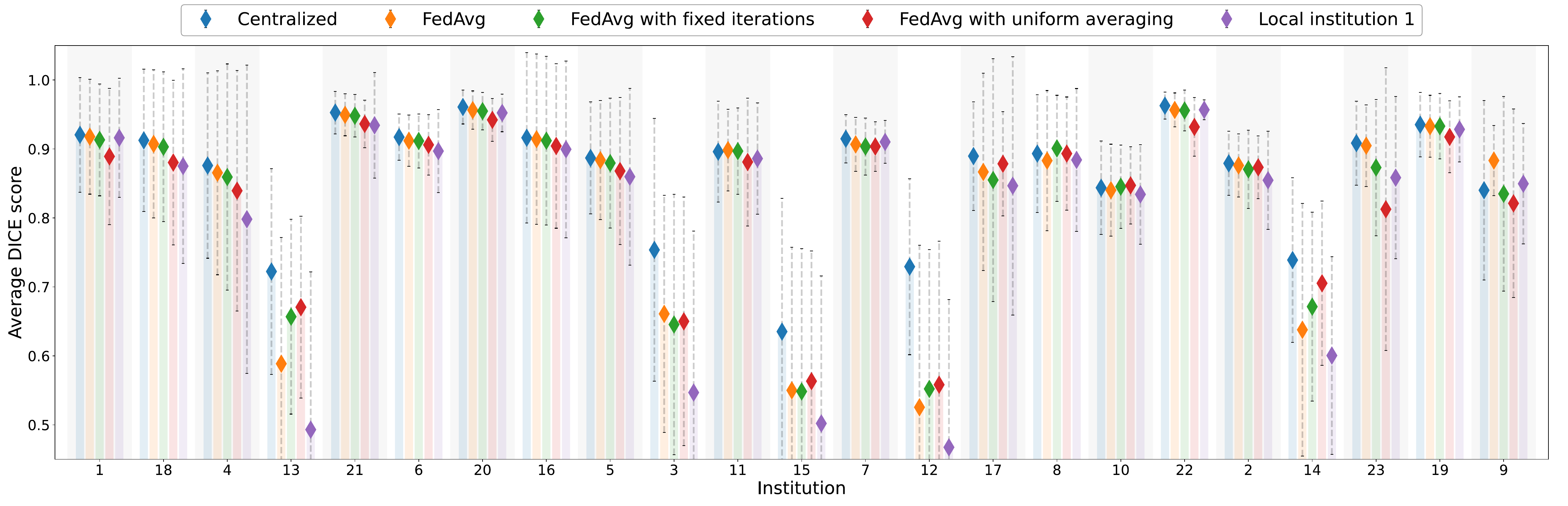}
\caption{Average Dice scores for \textit{local}, \textit{centralized} and \textit{FedAvg} variations training in the challenge setup of FeTS2022 dataset per institution, in decreasing order of number of samples (c.f. Figure \ref{fig:tumor_type_skew_fets2022}). Institutions 17 to 9 (on the right) each own from 10 to 4 samples in total. Errors bars represent $\pm$ one standard deviation. \textit{Centralized} training (blue) is the upper baseline. Drops in performance are perceived for every variants of \textit{FedAvg} compared to \textit{Centralized}, with significantly larger ones for institutions 12, 13, 14, 15 and 3. \textit{FedAvg with fixed epochs} (orange) converges faster than other variants to a close match with \textit{Centralized} performances at the cost of larger gaps for previously cited institutions.}
\label{fig:fed_avg_inst}
\end{figure*}

We show in Table \ref{tab:results} dice scores and 95\% Hausdorff distances obtained by different variations of \textit{FedAvg} compared to the performance of \textit{Local institution 1} and \textit{Centralized} training, while Figure \ref{fig:fed_avg_inst} shows the average dice scores per institution. First, institution 1 is by itself capable of training a very decent model which however does not generalize well to the dataset of other institutions. Furthermore, centralizing the volumes of every institution does bring a significant improvement ($p \ll 10^{-10}$) of the segmentation accuracy (Dice and Hausdorff) on all of them, with still a slight improvement for samples of institution 1. We see in Figure \ref{fig:fed_avg_inst} however that the centralized model does not perform as well on institutions 3, 12, 13, 14 and 15 as on others.

\textit{FedAvg} with weighted aggregation and  a fixed amount of local epochs per round enables training of a model with very close performances to the \textit{centralized} one, with still a significant gap ($p \ll 10^{-10}$) of one dice point and a slight edge in 95\% Hausdorff distance. Performances on institutions 3, 12, 13, 14, and 15 are however significantly worse than in the centralized case (Figure \ref{fig:fed_avg_inst}) ($p= 0.04, 5\times10^{-4}, 5\times10^{-6}, 0.015$ and $0.02$ respectively). \textit{FedAvg w/ fixed local epochs} does focus on the performance of large and representative institutions, while institutions that appear to own data further away from the average distribution are clearly penalized compared to centralized training.

Compared to the previous variant of \textit{FedAvg}, using a uniform averaging or a fixed amount of iterations per communication round slows down the convergence, with significantly lower average performance ($p \ll 10^{-10}$) both in terms of dice score and Hausdorff distance. The performance loss compared to \textit{centralized} on institution 13 is however less important. This recovers theory since using both a weighted aggregation and one local epoch per round ``doubly'' increases the importance to institutions owning a large number of samples, orienting the optimization toward their local objective \cite{wang_tackling_2020}. This shows however that a compromise must be made: giving this much weight to institutions 1 and 18 does enable them to "drive" the federation faster toward an efficient global model, at the cost of limiting its performance on institutions owning atypical samples.

We use in the remainder of this article \textit{FedAvg w/ fixed local epochs} as the baseline for the comparison to other SOTA methods.

~
\begin{table*}[h!]\small
\centering
\caption{\label{tab:results} Dice scores and 95\% Hausdorff distances (mm) obtained by \textit{local}, \textit{centralized} and SOTA federated learning algorithms on FeTS2022 (average $\pm$ std over samples of all folds)}
\begin{tabular}{|l|l|l|l|l|l|}
\hline
\textbf{Algorithm}       & \textbf{Mean DICE}  & \textbf{DICE WT} & \textbf{DICE TC} & \textbf{DICE ET} & \textbf{Mean 95HD}\\

\hline\hline
\textbf{Centralized} & \textbf{0.903 $\pm$ 0.109}&	\textbf{0.929 $\pm$	0.072}&	\textbf{0.907 $\pm$	0.154}&	\textbf{0.873 $\pm$	0.177} & \textbf{4.565 $\pm$ 6.561}\\ \hline
Local training institution 1 & 0.871 $\pm$	0.159&	0.903 $\pm$	0.112&	0.868 $\pm$	0.221&	0.841 $\pm$	0.210 & 7.342 $\pm$	11.403\\

\hline\hline
\rowcolor[RGB]{210, 210, 210} \multicolumn{6}{|c|}{FedAvg variants} \\\hline
FedAvg w/ fixed local epochs & 0.891 $\pm$	0.130&	0.924 $\pm$	0.079&	0.892 $\pm$	0.187&	0.858 $\pm$	0.195 & 4.742 $\pm$	6.270\\ \hline
FedAvg w/ fixed local iterations & 0.889 $\pm$	0.124&	0.921 $\pm$	0.081&	0.895 $\pm$	0.167&	0.852 $\pm$	0.196 & 5.054 $\pm$	6.949\\ \hline
%FedAvg w/ fixed local iterations & 0.871 $\pm$	0.138&	0.908 $\pm$	0.087&	0.872 $\pm$	0.190&	0.833 $\pm$	0.206 & 6.519 $\pm$	9.779\\ \hline
FedAvg w/ uniform averaging & 0.871 $\pm$	0.128&	0.906 $\pm$	0.087&	0.876 $\pm$	0.173&	0.831 $\pm$	0.203 & 6.541 $\pm$	9.306\\ 

\hline\hline
\rowcolor[RGB]{210, 210, 210} \multicolumn{6}{|c|}{Global federated methods} \\\hline
FedNova & 0.877 $\pm$	0.124&	0.910 $\pm$	0.083&	0.883 $\pm$	0.172&	0.839 $\pm$	0.195&	5.824 $\pm$	7.532\\\hline
FedAdam & 0.886 $\pm$	0.135&	0.922 $\pm$	0.082&	0.888 $\pm$	0.180&	0.847 $\pm$	0.209 & 6.026 $\pm$	10.919\\ \hline
\textit{SCAFFOLD} & \textit{0.897 $\pm$	0.113}	&	\textit{0.925 $\pm$ 0.079}& \textit{0.904 $\pm$ 0.154}&	\textit{0.863 $\pm$	0.187} & \textit{4.839 $\pm$	7.140}\\ \hline
q-FedAvg q=1.0 & 0.803 $\pm$	0.165&	0.861 $\pm$	0.137&	0.781 $\pm$	0.223&	0.768 $\pm$	0.225 & 12.994 $\pm$ 18.027\\\hline
FedPIDAvg & 0.891 $\pm$ 0.122  & 0.921 $\pm$	0.081 & 0.896 $\pm$ 0.172&	0.857 $\pm$	0.192 &	5.163 $\pm$	7.141\\

\hline\hline\rowcolor[RGB]{210, 210, 210} \multicolumn{6}{|c|}{Personalized and clustered federated methods} \\\hline
Local finetuning & 0.895 $\pm$	0.120&	0.924 $\pm$	0.079&	0.900 $\pm$	0.164&	0.861 $\pm$	0.194 & 5.114 $\pm$	7.233\\\hline
Ditto & 0.894 $\pm$	0.123&	0.924 $\pm$	0.082&	0.900 $\pm$	0.164&	0.859 $\pm$	0.197 & 5.123 $\pm$	7.379\\ \hline
FedPer & 0.886 $\pm$	0.135&	0.922 $\pm$	0.085&	0.894 $\pm$	0.173&	0.842 $\pm$	0.224 & 6.289 $\pm$ 11.335 \\ \hline
LG-FedAvg &  0.881 $\pm$	0.138&	0.914 $\pm$	0.094&	0.887 $\pm$	0.176&	0.841 $\pm$	0.214 & 6.987 $\pm$	11.776\\ \hline
CFL 2 clusters & 0.891 $\pm$	0.129&	0.923 $\pm$	0.081&	0.890 $\pm$	0.187&	0.859 $\pm$	0.194 & 5.114 $\pm$	7.780\\ \hline
\textit{Prior CFL HGG/LGG}& \textit{0.896 $\pm$	0.115}&	\textit{0.925 $\pm$	0.080}&	\textit{0.901 $\pm$	0.162}&	\textit{0.863 $\pm$	0.185} & \textit{4.753 $\pm$	6.568}\\

\hline\hline
\rowcolor[RGB]{210, 210, 210} \multicolumn{6}{|c|}{IID setup} \\\hline
FedAvg IID 23 clients 300 rounds & 0.878 $\pm$	0.126&	0.912 $\pm$	0.087&	0.885 $\pm$	0.165&	0.837 $\pm$	0.203 & 6.120 $\pm$	8.852\\ \hline
FedAvg IID 10 clients 300 rounds & 0.888 $\pm$	0.122&	0.920 $\pm$	0.079&	0.895 $\pm$	0.164&	0.849 $\pm$	0.198 & 5.372 $\pm$	7.853\\ \hline
FedAvg IID 5 clients 300 rounds & 0.897 $\pm$	0.111&	0.925 $\pm$	0.075&	0.903 $\pm$	0.158&	0.864 $\pm$	0.183 & 4.700 $\pm$	6.713\\\hline

\end{tabular}
\end{table*}

\subsubsection{Global solutions}
\begin{figure*}[!t]
\centering
\includegraphics[scale=.246]{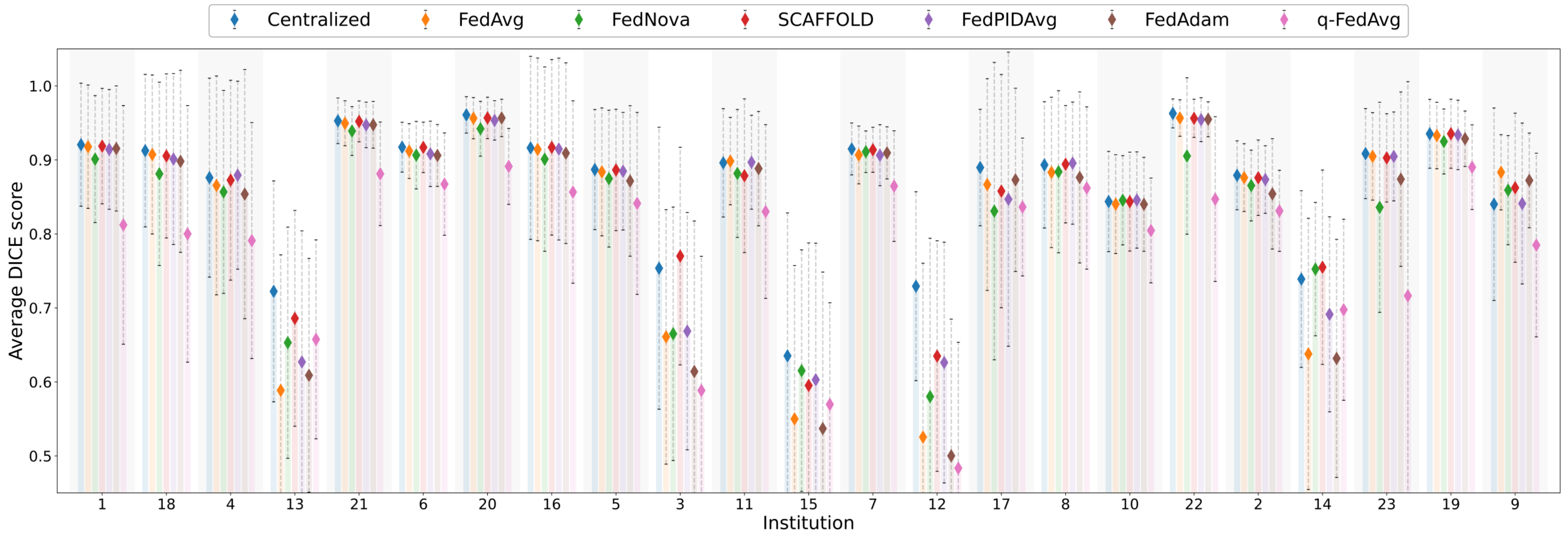}
\caption{Average Dice scores of \textit{centralized} and state-of-the-art global federated training in the challenge setup of FeTS2022 dataset per institution, in decreasing order of number of samples (c.f. Figure \ref{fig:tumor_type_skew_fets2022}). Institutions 17 to 9 each own from 10 to 4 total samples.  Errors bars represent $\pm$ one standard deviation.  Only \textit{SCAFFOLD} (red) improves upon the federated baseline \textit{FedAvg} (orange) by mitigating the drop in performance for institutions 12, 13, 14, 15 and 3 while maintaining it on others. \textit{q-FedAvg} (pink) and \textit{FedNova} (green) homogenizes DICE scores across institutions but decrease the overall performance.}
\label{fig:global_inst}
\end{figure*}

We show in Table \ref{tab:results} dice scores and 95\% Hausdorff distances obtained by SOTA global solutions compared to the baselines of \textit{Institution 1 local}, \textit{Centralized} and \textit{FedAvg} training, while Figure \ref{fig:global_inst} shows the average dice scores per institution.

First, \textit{FedAdam} does not improve the convergence of training over standard \textit{FedAvg}. To limit the computational cost and ensure the fairness of comparison to other methods, we limited the size of the grid search over the hyperparameters of this method, with $\beta_1, \beta_2, \tau$ fixed following the recommendations of \cite{reddi_adaptive_2022} and two learning rates remaining to tune (c.f. Table \ref{tab:hyperparameters_fets2022}). Better performances and convergence might be obtainable with a better choice of hyperparameters, but it is simply unrealistic in a federated learning context.

\textit{SCAFFOLD} on the other hand does provide a significant gain over \textit{FedAvg} in dice scores ($p = 0.03$), while remaining close in Hausdorff distance. We show in Figure \ref{fig:global_inst} that the performance gain can be perceived for most institutions, and in particular for 3, 12, 13, 14 and 15 ($p = 0.02, 0.09, 8\times10^{-6}, 0.016$ and $0.07$). In other words, the optimization and performance distribution experienced when using \textit{SCAFFOLD} is much closer to \textit{centralized} training than with \textit{FedAvg}, at the cost of twice larger communications. We can only recommend using this optimizer for this task if the network infrastructure enables it.

In the same fashion as \textit{SCAFFOLD}, \textit{FedNova} tends to improve the average dice scores for institutions 12, 13, 14 and 15, but significantly reduces the average dice score across all institutions ($p \ll 10^{-10}$).

\textit{q-FedAvg} does provide a fairer dice score distribution, while however bringing the performance of every other institution significantly down ($p \ll 10^{-10}$).

\textit{FedPIDAvg} gives similar performance to \textit{FedAvg}, while giving a smoother validation loss. The weighting dynamics of this method are very interesting for this task. The large institutions are given higher weights in the first rounds of training as they bring their local validation loss down way faster than others, promoting faster initial convergence. In the later rounds of training, more weight is given to other institutions which continue improving their loss, which justifies a better performance on institutions 12, 13, 14 and 15 than standard \textit{FedAvg}. However, similarly to \textit{FedAdam}, it might suffer from fixed $\alpha, \beta$ and $\gamma$ hyperparameters following \cite{machler_fedpidavg_2023}, as well as the randomness of local validations with a very small amount of samples.
~

\subsubsection{Personalized and hybrid solutions}
\begin{figure*}[!t]
\centering
\includegraphics[scale=.23]{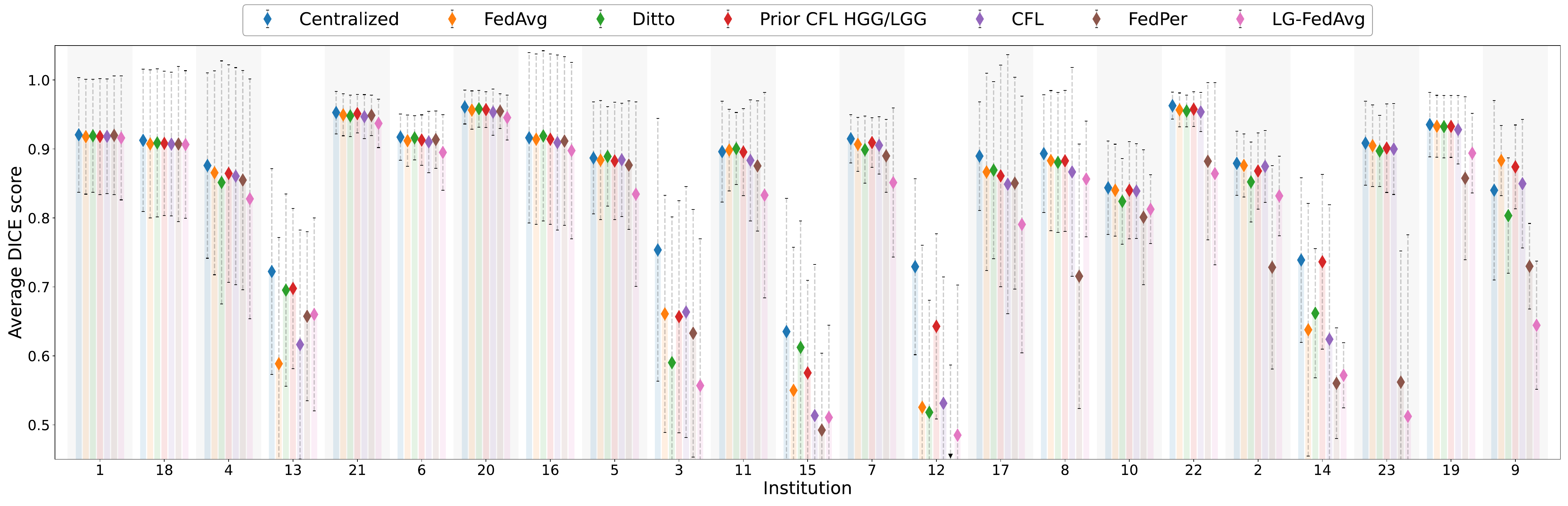}
\caption{Average Dice scores of \textit{centralized} and state-of-the-art personalized and hybrid federated training in the challenge setup of FeTS2022 dataset per institution, in decreasing order of number of samples (c.f. Figure \ref{fig:tumor_type_skew_fets2022}). Institutions 17 to 9 each own from 10 to 4 total samples.  Errors bars represent $\pm$ one standard deviation.  \textit{Local finetuning} was removed from this graph as it gives very similar DICE distribution as \textit{Ditto}. These two methods can improve the performance on large enough institutions like 13. The performance of partial model-sharing methods \textit{LG-FedAvg} and \textit{FedPer} collapses for very small institutions. \textit{Prior CFL HGG/LGG} mitigates the drop in performance of \textit{FedAvg} on institutions 12 to 15.}
\label{fig:perso_inst}
\end{figure*}

We show in Table \ref{tab:results} dice scores and 95\% Hausdorff distances obtained by SOTA personalized and hybrid solutions compared to the baselines of \textit{Institution 1 local}, \textit{Centralized} and \textit{FedAvg} training, while Figure \ref{fig:perso_inst} shows the average dice scores per institution.

We show first that \textit{local finetuning} does bring a significant performance improvement in dice score ($p = 10^{-8}$) while slightly increasing the average Hausdorff distance. We can note that overall, segmentation performance of the whole tumor (WT) is made a little bit more irregular with a slightly larger Hausdorff distance, but a clear improvement is perceived on the tumor core label ($p = 3\times10^{-4}$) . Institution-wise, we can see that \textit{local finetuning} does improve the performance for every institution owning a large number of samples, especially for institution 13 for which it enables to limit the performance gap between \textit{centralized} training and \textit{FedAvg} ($p = 0.0013$). Benefits become unclear for smaller institutions: finetuning a model with more than 22M parameters on only a few samples can definitely lead to overfitting. Even when selecting the best model based on validation performance, \textit{local finetuning} is not adapted to very small institutions. The regularization in \textit{Ditto} does not enable solving the problems of \textit{local finetuning}, giving very similar performances.

The explored partial model-sharing methods clearly decrease the performance of the model for most institutions ($p = 0.78$ and $10^{-11}$ for \textit{FedPer} and \textit{LG-FedAvg} respectively when including all samples, and $p = 10^{-5}$ and $10^{-17}$ when taking out samples from institutions 1 and 18 from the tests). They are also quite unadapted to the small number of samples of most institutions. The fewer samples an institution has the more the performance decreases compared to \textit{FedAvg}, which is coherent with the fact that they have to be able to train from scratch several layers of the network. We can note an interesting difference between \textit{FedPer} and \textit{LG-FedAvg}. The latter gives decent performance only on institutions 1 and 18 owning a large amount of data, equivalent to the local model they can obtain alone, with drastically decreased performances for institutions owning 40 samples or fewer. The former still gives decent performances for institutions 5, 6, 16, 20 and 21 which owns between 22 and 35 samples each. \textbf{In that sense, the federation of the first layers of the model seems to be very important for better performance}, or they at least require to be trained on a larger amount of samples than the last layers to be meaningful.

This section's results motivate the study of clustering-based personalization methods for this task, potentially addressing issues caused by small local sample sizes. We show however that the pure gradient similarity-based clustered method \textit{CFL} does not provide a significant performance gain over other methods, even compared to classical \textit{FedAvg}. While we remarked that institutions 12, 13 and 14 were always clustered together, the computed clusters vary considerably from one fold to another. We could not find a concrete link between institutions clustered together for each fold. This method seems to suffer from patch training and the extremely unbalance setup. We show however that applying \textit{FedAvg} to clusters of institutions that are defined on prior knowledge of the tumor grade does bring a significant performance gain  on institutions 12 to 15 which were isolated together ($p = 10^{-5}$). This however requires the communication of potentially sensitive information on the data distribution of each institution which may be acceptable or not. 
Overall, this clearly motivates the study of hybrid personalization in the context of a low amount of samples for most institutions.

~

\subsection{Cost estimations on FeTS2022 original partitioning}

We show in Table \ref{tab:complete_tab_cost} cost estimations of the different algorithms applied on the complete FeTS2022 federated dataset. 

Comparing \textit{FedAvg w/ fixed epochs} and \textit{w/ fixed iterations} is highly dependent on the hypotheses on the cost of computations and communications. Our choice of choosing a realistic lower bound on the communication costs while neglecting stragglers effect clearly advantages \textit{FedAvg w/ fixed iterations}, but this algorithm still slightly, but significantly, underperforms compared to \textit{FedAvg w/ fixed epochs}, although being less biased. It also consumes much more energy since it requires around three times more total SGD steps than \textit{FedAvg w/ fixed epochs}.
\textit{SCAFFOLD} does necessitate two times more communications than \textit{FedAvg}. This increase appears relatively negligible in the simulated training time.
We allowed for slightly more training time for \textit{Local finetuning} and \textit{Ditto}, as finetuning-based methods are generally used after global training ended. These additional local steps do not require synchronization between institutions anymore, with the main improvement being brought by these methods generally in the first one to five local epochs before overfitting.

\begin{table*}[h!]\small
    \centering
    \caption{\label{tab:complete_tab_cost} Cost evaluation of each algorithm of the benchmark in the challenge setup.}
    \begin{tabular}{|c|c|c|c|c|}
        \hline
        \multirow{2}{*}{\textbf{Algorithm}} & \textbf{Total}  & \textbf{Total maximum}  & \textbf{Total communications}  & \textbf{Estimated}  \\
         & \textbf{SGD steps} & \textbf{parallel SGD steps} & \textbf{(number of floats $\times10^9$)} & \textbf{training time (h)} \\\hline\hline
         \textit{Local institution 1} & 24600 & 24600 & 0 & 18.2\\\hline
         \textit{Centralized} & 59400 & 59400 & 0 & 44.6\\\hline\hline
         \rowcolor[RGB]{210, 210, 210} \multicolumn{5}{|c|}{FedAvg variants} \\\hline
         \textit{FedAvg w/ fixed local}  & \multirow{2}{*}{59400} & \multirow{2}{*}{24600} & \multirow{2}{*}{13.5} & \multirow{2}{*}{19.1}\\
         \textit{epochs} & & & & \\\hline
         \textit{FedAvg w/ fixed local} & \multirow{2}{*}{165600} & \multirow{2}{*}{7200} & \multirow{2}{*}{32.5} & \multirow{2}{*}{19.1}\\
         \textit{iterations} & & & & \\\hline\hline
         \rowcolor[RGB]{210, 210, 210} \multicolumn{5}{|c|}{Global federated methods} \\\hline
         \textit{FedNova}  & \multirow{4}{*}{59400} & \multirow{4}{*}{24600} & \multirow{4}{*}{13.5} & \multirow{4}{*}{19.1}\\
         \textit{FedAdam} & & & & \\
         \textit{FedPIDAvg} & & & & \\
         \textit{q-FedAvg} & & & & \\\hline
         \textit{SCAFFOLD} & 59400 & 24600 & 27.0 & 19.9\\\hline\hline
         \rowcolor[RGB]{210, 210, 210} \multicolumn{5}{|c|}{Personalized and clustered federated methods} \\\hline
         \textit{Local finetuning} & \multirow{2}{*}{65340} & \multirow{2}{*}{27060} & \multirow{2}{*}{13.5} & \multirow{2}{*}{20.9}\\
         \textit{Ditto} & & & & \\\hline
         \textit{FedPer} & 59400 & 24600 & 13.5 & 19.1 \\\hline
         \textit{LG-FedAvg} & 59400 & 24600 & 13.4 & 19.1\\\hline
         \textit{CFL} & \multirow{2}{*}{59400} & \multirow{2}{*}{24600} & \multirow{2}{*}{13.5} & \multirow{2}{*}{19.1}\\
         \textit{Prior CFL HGG/LGG} & & & & \\\hline 
    \end{tabular}
\end{table*}

\subsection{IID setup}
\begin{figure*}[!t]
\centering
\includegraphics[scale=.245]{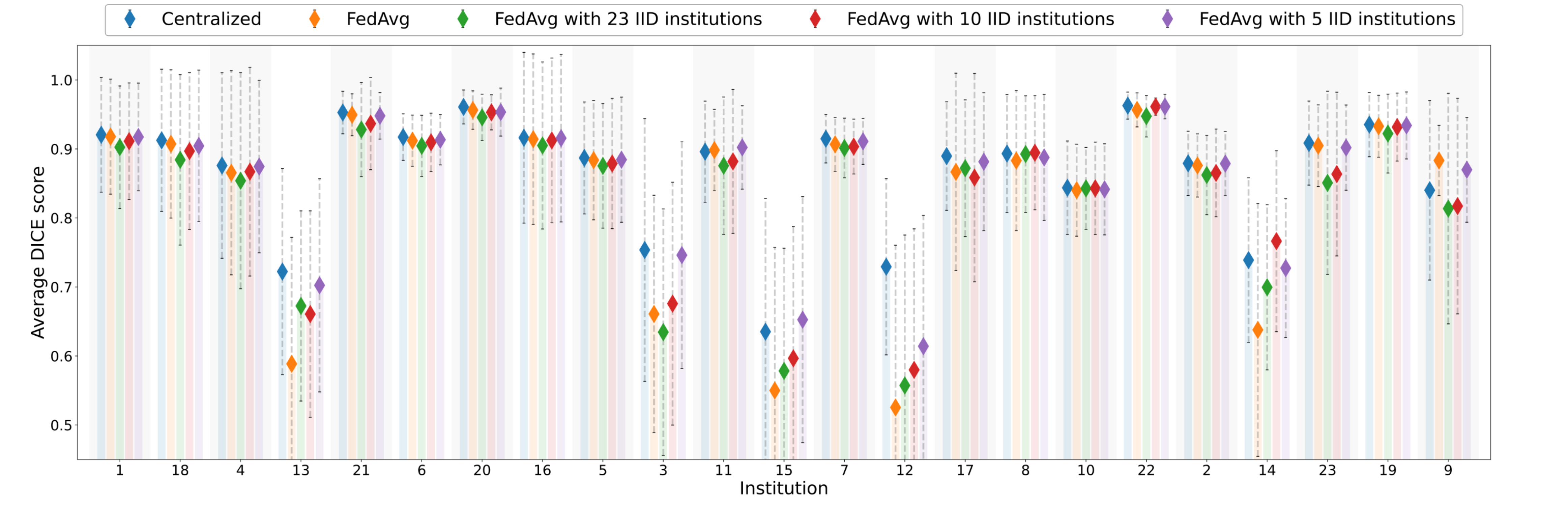}
\caption{Average Dice scores of \textit{centralized} and \textit{FedAvg} training in the challenge and \textit{IID} setups of FeTS2022 dataset per institution, in decreasing order of number of test samples (c.f. Figure \ref{fig:tumor_type_skew_fets2022}).  Errors bars represent $\pm$ one standard deviation.  An IID setup limits the drop in performance of \textit{FedAvg} on institutions 12, 13, 14, 15 and 3.}
\label{fig:iid_inst}
\end{figure*}

We show in Table \ref{tab:results} dice scores and 95\% Hausdorff distances obtained by applying \textit{FedAvg} in the IID setup described in Section \ref{sec:iid_case} compared to the baselines of \textit{Institution 1 local}, \textit{Centralized} and \textit{FedAvg} training, while Figure \ref{fig:iid_inst} shows the average dice scores per institution. Validation and test splits were preserved for every fold, only the training samples were redistributed such that performances between the original and IID setup remain comparable.

Interestingly, we show that applying \textit{FedAvg} after redistributing the training samples in an i.i.d and balanced fashion to 23 synthetic institutions leads to a significant decrease in performance compared to the original data distribution ($p \ll 10^{-10}$), both in terms of dice score and Hausdorff distance. In that sense, this again motivates the fact that institutions 1 and 18 owning the vast majority of samples in the original FeTS2022 setup drives the federated training toward a decent global model, even in the presence of statistical heterogeneity between local data distributions. Performances on institutions 12 to 15 are on the contrary slightly larger than \textit{FedAvg}, since in the IID setup \textit{FedAvg} does not prioritize the performance of samples owned by institutions 1 and 18.

Performance increases when we decrease the number of synthetic institutions, with five synthetic institutions being small enough to enable \textit{FedAvg} in the IID setup to outperform \textit{FedAvg} in the original setup on average, without statistically significant differences between medians ($p = 0.53$). In this configuration, each of the five synthetic institutions owns around 160 samples, much less than institutions 1 and 18 in the original setup. While institutions 1 and 18 perform a larger number of updates, enabling them to drive  \textit{FedAvg} training in the challenge setup, a lower amount per institution is required in the IID setup to match its performance. We hypothesize that this convergence speed gap can be caused by the statistical heterogeneity in the original distribution.

~

\subsection{Limited data setup}
We show in Table \ref{tab:small} dice scores and 95\% Hausdorff distances obtained by applying \textit{FedAvg} in the limited data setup described in Section \ref{sec:limited_case} compared to the baselines of \textit{Institution 1 local}, \textit{Centralized} and \textit{FedAvg} training, with average Dice scores per institutions in Figure \ref{figknown_tumor_inst} of Appendix \ref{app:kown_tumor}.

The performance is overall much worse than in the previous cases, due to the drastic reduction of the total sample size. We recover in this setup the same difference between \textit{FedAvg} and \textit{Centralized} training, with a relatively small but significant gap ($p = 10^{-10}$). \textit{SCAFFOLD} in this setup however does not improve the performance over \textit{FedAvg}, it is actually the opposite in median dice score ($p = 0.002$). \textit{HGG/LGG prior CFL} does not either significantly improve the segmentation performance over \textit{FedAvg} in this setup ($p = 0.13$), although a slight gain might be perceivable in Tumor Core. The overall amount of data seems too low for clustered finetuning to be useful.  The performance on institutions 12, 13, 14, 15 and 3 remains worse than others for every optimizer, including \textit{Centralized} training. Their samples tend to be harder to be segmented by a U-Net, with any optimization seeming to prioritize the performance of the predominant HGG samples.

~
\begin{table*}[h!]\small
\centering
\caption{\label{tab:small}Dice scores and 95\% Hausdorff distances (mm) obtained by \textit{centralized} and selected state-of-the-art federated training in the \textit{limited setup} of FeTS2022 dataset (average $\pm$ std over samples of all folds)}
\begin{tabular}{|l|l|l|l|l|l|}
\hline
\textbf{Algorithm}       & \textbf{Mean DICE}  & \textbf{DICE WT} & \textbf{DICE TC} & \textbf{DICE ET} & \textbf{Mean 95HD}\\ \hline\hline
\textbf{Centralized} & 0.815 $\pm$	0.161&	0.904 $\pm$	0.098&	0.822 $\pm$	0.188&	0.719 $\pm$	0.308 & 7.823 $\pm$	11.962\\ \hline\hline
FedAvg w/ fixed local epochs & 0.806 $\pm$	0.164&	0.896 $\pm$	0.099&	0.812 $\pm$	0.187&	0.711 $\pm$	0.307 & 8.715 $\pm$	11.572\\ \hline\hline
SCAFFOLD & 0.805 $\pm$	0.160	&	0.898 $\pm$	0.085&0.811 $\pm$	0.185&	0.707 $\pm$	0.304&	7.995 $\pm$	10.613\\\hline
Local finetuning & 0.806 $\pm$	0.165&	0.891 $\pm$	0.105&	0.808 $\pm$	0.196&	0.718 $\pm$	0.305 & 9.332 $\pm$	12.678\\\hline
Prior CFL HGG/LGG & 0.807 $\pm$	0.162&	0.897 $\pm$	0.092&	0.819 $\pm$	0.162&	0.706 $\pm$	0.305 & 8.928 $\pm$	12.651\\\hline
\end{tabular}
\end{table*}

\section{Discussion}\label{sec:discussion}

% Rajouter une discussion en premier, "guidelines" plus haut niveau de description des résultats.
% Ici des images de cerveau, 
% Puis le choix des méthodes (numéro 2).
% Partie limitation en dernier.

\subsection{Results synthesis and takeaways}

We proposed an extensive benchmark of SOTA federated methods on different data partitionings of the FeTS2022 dataset. In the challenge setup, standard \textit{FedAvg} with a fixed amount of local epochs and weighted averaging performs very closely to \textit{Centralized} training. The difference in performance between variants of \textit{FedAvg} is clear, we think that the mismatch between practice and theory could motivate the analysis of local epochs-based federated optimizers theoretically. From all the global methods experimented with, \textit{SCAFFOLD} is the only one significantly improving the average segmentation performance compared to baselines. It seems like this federated optimizer behaves much more like \textit{Centralized} training than \textit{FedAvg}. This however does not hold in the limited setup. While personalization methods were hampered by the overall low amount of samples per institution, \textit{local finetuning} and \textit{Ditto} could still provide small benefits for large enough institutions. Considering the data distribution of the dataset, with LGG and HGG samples presenting dissimilar label distributions, clustered FL based on this prior knowledge significantly improves the segmentation performance in the challenge. We think that this class of methods is of particular interest in cross-silo medical image analysis tasks  with enough data .

\begin{figure*}[!t]
\centering
\includegraphics[scale=.058]{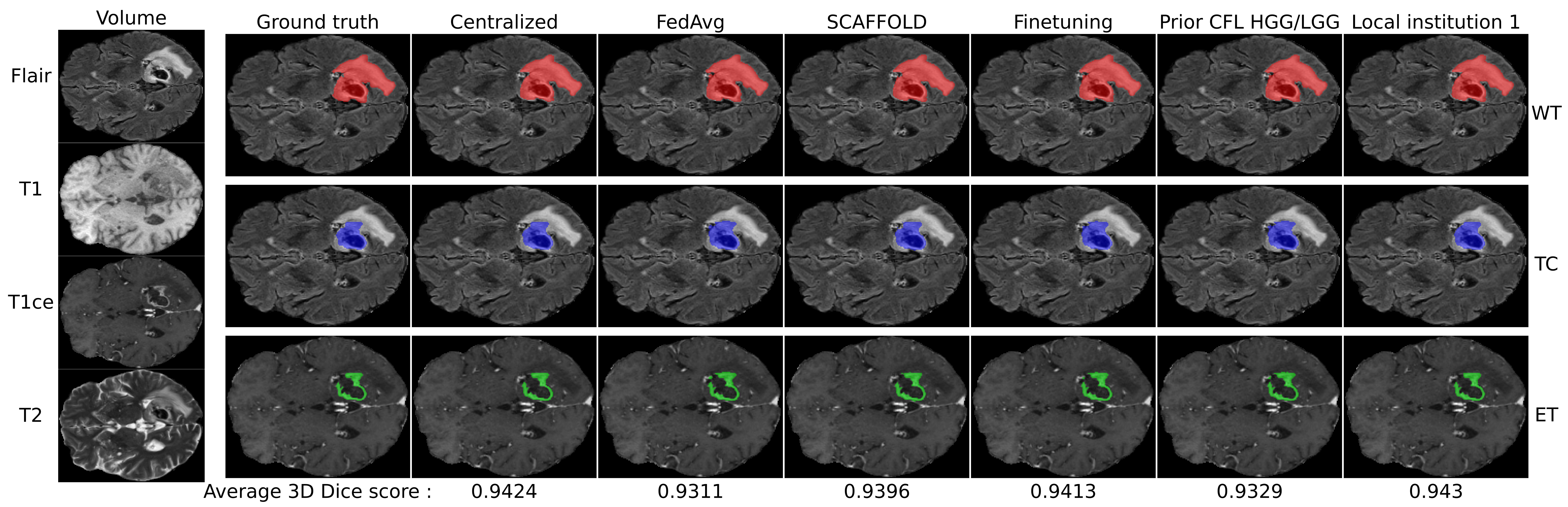}
\caption{Qualitative comparison on example axial slice from patient n°00440 from institution 1, representative of most samples of this institution. First column shows, from top to bottom, the original FLAIR, T1, T1ce and T2 MRI images. On the second column are ground truth segmentations for this slice, and on next columns are the segmentations obtained by the 3D UNet model trained with the different FL algorithms. The three rows correspond to the three segmentation labels (from top to bottom, Tumor Core (TC), Whole tumor (WT) and Enhancing Tumor (ET)). \textit{FedAvg} and other federated methods give a well-performing model on the majority of the samples of the dataset.}
\label{fig:example_inst_1}
\end{figure*}

\begin{figure*}[!t]
\centering
\includegraphics[scale=.057]{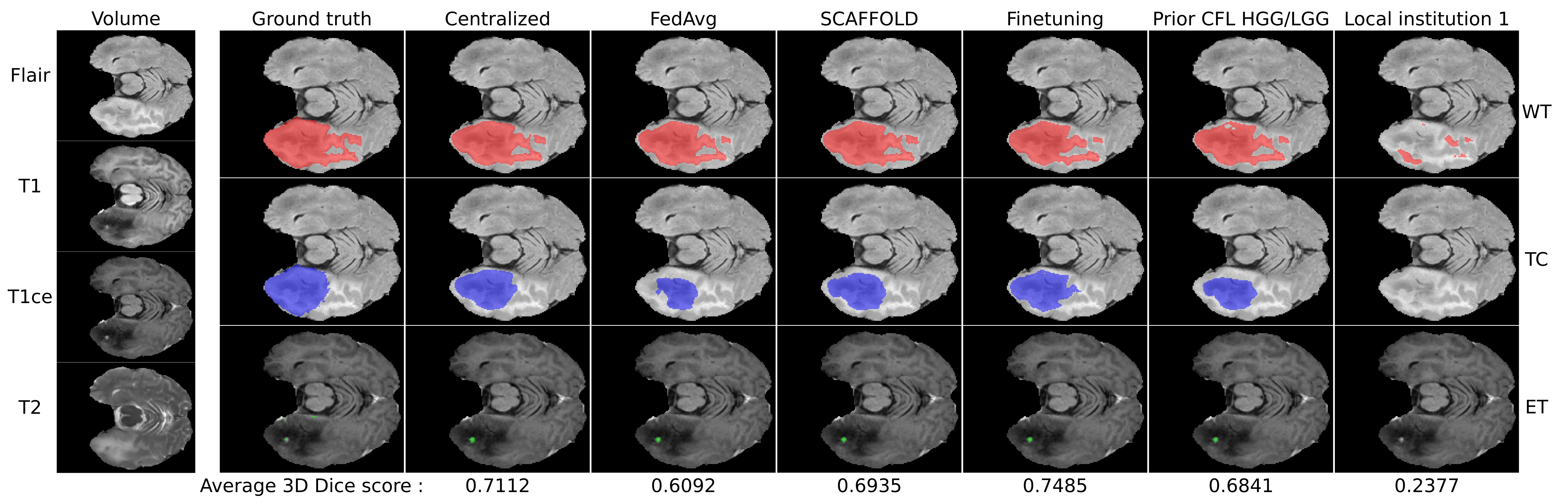}
\caption{Qualitative comparison on axial slice from patient n°01514 from institution 13. The figure is structured in the same way as Figure \ref{fig:example_inst_1}. For atypical samples from institutions 12, 13, 14, 15 and 3, the choice of federated learning algorithm has an important impact on the model's segmentation.}
\label{fig:example_inst_13}
\end{figure*}

For a more qualitative comparison, we provide in Figure \ref{fig:example_inst_1} and \ref{fig:example_inst_13} the ground truth and segmentation maps output by models trained with various methods on axial slices of two chosen samples. The first one on Figure \ref{fig:example_inst_1} is a sample from Institution 1 segmented accurately by most methods, representative of a large number of volumes in the dataset. The second one on Figure \ref{fig:example_inst_13} is an LGG sample from Institution 13 on which models underperform, representative of most samples from institutions 12 to 15. We see in Figure \ref{fig:example_inst_1} that the segmentations are very similar. A 0.5 to 1.0 \%  Dice score difference does not change fundamentally their shape, although some mistakes in the ET label appear in the segmentation made the model trained with \textit{FedAvg} compared to \textit{Centralized}. On the contrary, large differences in segmentation results can be seen in figure \ref{fig:example_inst_13}, where a large part of the TC label was omitted by the model trained with \textit{FedAvg} compared to baselines. From both of these examples, \textit{SCAFFOLD} seems to behave much more like centralized training, even when comparing shapes of the obtained segmentations. 

It is intriguing to see how good the performance of standard \textit{FedAvg} is compared to centralized training in the challenge setup. Compared to the IID setup with 23 institutions, \textit{FedAvg} in the challenge setup requires fewer communication rounds to achieve close to centralized performance. We think it is mainly due to institutions 1 and 18 each "centralizing" a large number of samples, driving federated training with large updates. While unbalanced setups could have a priori been seen as a problem, extreme unbalance might actually be exploitable to accelerate federated training.  Data-sharing strategies were already proposed in the literature to accelerate federated training, sharing only a fraction of the total data to the server \cite{zhao_federated_2018}. We show here that the presence of a large institution, potentially built as a coalition of entities inside the federation, with a sufficiently representative set of samples can be of interest in cross-silo setups. We do not know what type and level of heterogeneity such an extreme unbalance would be robust to, but the goal of newly defined optimizers in this setup becomes to limit the bias induced by this unbalance toward the large institutions. In essence, this might motivate the complementarity between the deployment of federated learning and the construction of large public medical image data centres: these large public datasets could be leveraged at server-side to become the institution driving the others.

 Finally, while we tried to roughly balance estimated training times among algorithms, it was based on debatable hypotheses on computation and communication costs. The actual training time of algorithms distributing computation and communication in very different ways such as \textit{FedAvg w/ fixed local epochs} and \textit{fixed local iterations} would be highly dependent on the actual federated infrastructure in practice. The reported estimated training times in Table \ref{tab:complete_tab_cost} are comparable between each other and only serve as a way to get a fair benchmark, but would not be representative of real federated training times as we proposed a very simplistic estimation model (we do not consider random stragglers for example).

\subsection{Choice of SOTA methods}
%S Je changeerais éventuellement le titre en "Other SOTA methods" 
We emphasize in this section that the choice of SOTA methods to include in the benchmark was not random. We covered as many lines of research as possible matching the task problems in Section \ref{sec:related_works} and tried to include in the benchmark one method per "class". Difficulties and benefits from each method can in some cases be translatable to methods of the same class (for example, most partial model-sharing methods would be harmed by the very low amount of samples most institutions own). 

Moreover, we voluntarily did not consider some SOTA methods in this benchmark. Ensemble knowledge distillation-based methods such as \textit{FedDF} \cite{lin_ensemble_2021} require access to public or generated data server-side, which seemed impractical in our medical context. Moreover, most distillation-based methods require the communication of segmentation maps, which can end up being expensive relative to the size of a convolutional model. \textit{pFedHN} \cite{shamsian_personalized_2021} could not scale enough to be used in our case, the segmentation model to be generated by pFedHN's hypernetwork is too large.

We also focused on methods requiring close to the same amount of communication and computation as \textit{FedAvg}, for fair comparability but also simply due to computational constraints, up to a maximum of two times larger computational and communication cost compared to \textit{FedAvg}. We could not experiment with \textit{FedFOMO} \cite{zhang_personalized_2021}, requiring too many forward passes per communication round.

\subsection{Limitations of the proposed benchmark}
We focused in this study on the performance side of federated learning for brain tumor segmentation. While federated learning was initially introduced as a privacy-preserving framework, it was quickly shown that this setup can still be attacked, potentially leaking information on local data to the server or an individual attacker \cite{rodriguez_barroso_survey_2023}. We do not cover in this study the combination of SOTA federated learning methods with privacy-preserving techniques such as differential privacy \cite{dwork_algorithmic_2013} or heavier cryptographic primitives for computational reasons and leave this for future work. For the same reason, we could not incorporate communication optimization techniques such as gradient sparsification or quantization \cite{kairouz_advances_2021}.

We only use standard SGD as the local optimizer in every experiments. It was shown in a variety of works that local adaptivity, if not handled well, might lead to inconsistencies in training \cite{li_privacy_preserving_2019, wang_local_2021, jin_accelerated_2022}. As there is no consensus on how local momentum should be applied in a federated setup, for the "safety" of this benchmark and its consistency with the implemented SOTA methods such as \textit{SCAFFOLD}, we chose standard SGD as the local optimizer. The exploration of local optimizers such as SGD with different types of momentum or Adam is a valuable line of exploration left for future work.

Cross-silo federated learning involving health institutions requires them to invest in a technical pole, buy GPU(s) and set up a complex and secure network environment for this task alone. Assuming the full availability of each platform for federated training seems natural, justifying the assumption of full participation of each institution in each federated round.

In this context, centralized federated learning (with a central server orchestrating training) is more secure and demands less communication than fully-decentralized (peer-to-peer) learning on a complete communication graph while giving the exact same results. Indeed, each institution would behave exactly as a central server in this scenario. We thus focused exclusively on the centralized setup, letting the exploration of gossip algorithms on sparse communication graphs for other works.

 We only investigated generalization performance of trained models on institutions taking part in training and not on unseen institutions as in \cite{yuan_what_2022}. The choice of institutions to leave out of training for unseen tests would greatly impact the federated learning behaviour in our case, and performing a nested-cross-validation on this aspect was out of our computational limitations. This would be a great line of work, especially considering the recently published BraTS 2023 cluster of datasets \cite{adewole_brain_2023, kazerooni_brain_2024}.

Finally, our choice to report performance of \textit{Prior CFL HGG/LGG} was not aimed at advocating the use of this specific prior-based CFL method for this task. It showed the potential benefits of using domain knowledge in federated learning, that there exists at least one institution clustering of FeTS2022 institutions for which finetuning on each cluster gives a significant performance improvement, and thus that current SOTA gradient-based CFL methods are not completely able to capture the intricacies of the dataset, highlighting a need for further research. The practicality of clustering data as a function of the tumor grade can be questioned. It seems reasonable to assume the availability or the possibility of acquiring this information for training samples, even in other practical scenarios than FeTS. For just acquired test volumes however, this tumor grade information would most probably be lacking. It then opens a new question: how can we use in practice \textit{Prior CFL} on test samples for which we have very limited information.

\section{Conclusion}
We experimented in this work with a large variety of SOTA federated learning methods on the recently published FeTS2022 dataset and tried to isolate classes of federated methods showing potential on this task. We showed that both SOTA global, personalized and hybrid solutions can improve the segmentation performance compared to standard \textit{FedAvg}, with \textit{SCAFFOLD} and hybrid federated learning seeming the most adapted for cross-silo medical image segmentation tasks. We hope that this study can be helpful both for future methodological works as well as real-world application of federated learning on similar tasks.

\section*{Acknowledgement}
This work was partially supported by the Agence Nationale de la Recherche under grant ANR-20-THIA-0007 (IADoc@UdL). It was granted access to the HPC resources of IDRIS under the allocation 2023-AD011013327R1 made by GENCI.

%%Harvard
\bibliographystyle{model2-names.bst}\biboptions{authoryear}
\bibliography{refs}

\newpage

\appendix

\section{Hyperparameters search and selected values}\label{app:hyperparam}

\begin{table*}[h]\small
\caption{\label{tab:hyperparameters_fets2022} Hyperparameters search in the different FeTS2022 partitionings}
\centering
\begin{tabular}{|l|l|l|l|}
\hline
\textbf{Method}      & \textbf{Parameter} & \textbf{Search space} & \textbf{Selected value} \\ \hline\hline
\rowcolor[RGB]{210, 210, 210} \multicolumn{4}{|c|}{Challenge setup} \\\hline
\textit{Centralized, institution 1 local} & learning rate & $\{0.008, 0.02, 0.05, 0.1, 0.2\}$ & 0.1\\\hline
\textit{FedAvg and variations, SCAFFOLD, CFL} & learning rate & $\{0.02, 0.05, 0.1, 0.2, 0.4, 0.8\}$ & 0.4\\\hline
\textit{FedNova} & learning rate & $\{0.05, 0.1, 0.2, 0.4\}$ & 0.2\\\hline
\multirow{3}{*}{\textit{FedAdam}} & local learning rate & $\{0.01, 0.1, 1.0\}$ & 0.1\\\cline{2-4}
                                    & server learning rate & $\{0.0001, 0.001, 0.01, 0.1, 1.0\}$ & 0.001\\\cline{2-4}
                                    & $\beta_1, \beta_2, \tau$ & $\emptyset$ & 0.9, 0.999, $10^{-8}$\\\hline
\multirow{2}{*}{\textit{q-FedAvg}} & learning rate & $\{0.05, 0.1, 0.2, 0.4, 0.8\}$ & 0.4\\\cline{2-4}
                                & $q$ & $\{0.01, 0.1, 1.0, 2.0\}$ & 1.0\\\hline
\multirow{2}{*}{\textit{FedPIDAvg}} & learning rate & $\{0.05, 0.1, 0.2, 0.4\}$ & 0.4\\\cline{2-4}
                                & $\alpha, \beta, \gamma$ & $\emptyset$ & 0.45, 0.45, 0.1\\\hline
\textit{Local finetuning} & learning rate & $\{0.01, 0.02, 0.05, 0.1, 0.2, 0.4\}$ & per institution\\\hline
\multirow{2}{*}{\textit{Ditto}} & learning rate & $\{0.01, 0.04, 0.1, 0.4\}$ & \multirow{2}{*}{per institution}\\\cline{2-3}
                                & $\lambda$ & $\{0.001, 0.01, 0.1, 1.0\}$ & \\\hline
\multirow{2}{*}{\textit{FedPer}} & learning rate & $\{0.02, 0.05, 0.1, 0.2, 0.4\}$ & 0.4\\\cline{2-4}
                                & number of private layers & $\emptyset$ & 4\\\hline
\multirow{2}{*}{\textit{LG-FedAvg}} & learning rate & $\{0.02, 0.05, 0.1, 0.2, 0.4\}$ & 0.2\\\cline{2-4}
                                    & number of private layers & $\emptyset$ & 4\\\hline
\textit{HGG/LGG CFL} & learning rate & $\{0.02, 0.05, 0.1, 0.2\}$ & 0.1\\\hline\hline
\rowcolor[RGB]{210, 210, 210} \multicolumn{4}{|c|}{IID setup} \\\hline
\textit{FedAvg IID} & learning rate & $\emptyset$ & 0.4\\\hline\hline
\rowcolor[RGB]{210, 210, 210} \multicolumn{4}{|c|}{Limited setup} \\\hline
\textit{Centralized, institution 1 local} & learning rate & $\{0.05, 0.1, 0.3, 0.8\}$ & 0.1\\\hline
\textit{FedAvg and SCAFFOLD} & learning rate & $\{0.05, 0.1, 0.2, 0.4, 0.8\}$ & 0.8\\\hline
\textit{Local finetuning} & learning rate & $\{0.01, 0.02, 0.05, 0.1, 0.2, 0.4\}$ & per institution\\\hline
\textit{HGG/LGG CFL} & learning rate & $\{0.02, 0.05, 0.1, 0.2, 0.4\}$ & 0.2\\\hline
\end{tabular}
\end{table*}

\section{Additional figures of the limited data setup}\label{app:kown_tumor}

\begin{figure*}[h]
\centering
\includegraphics[scale=.245]{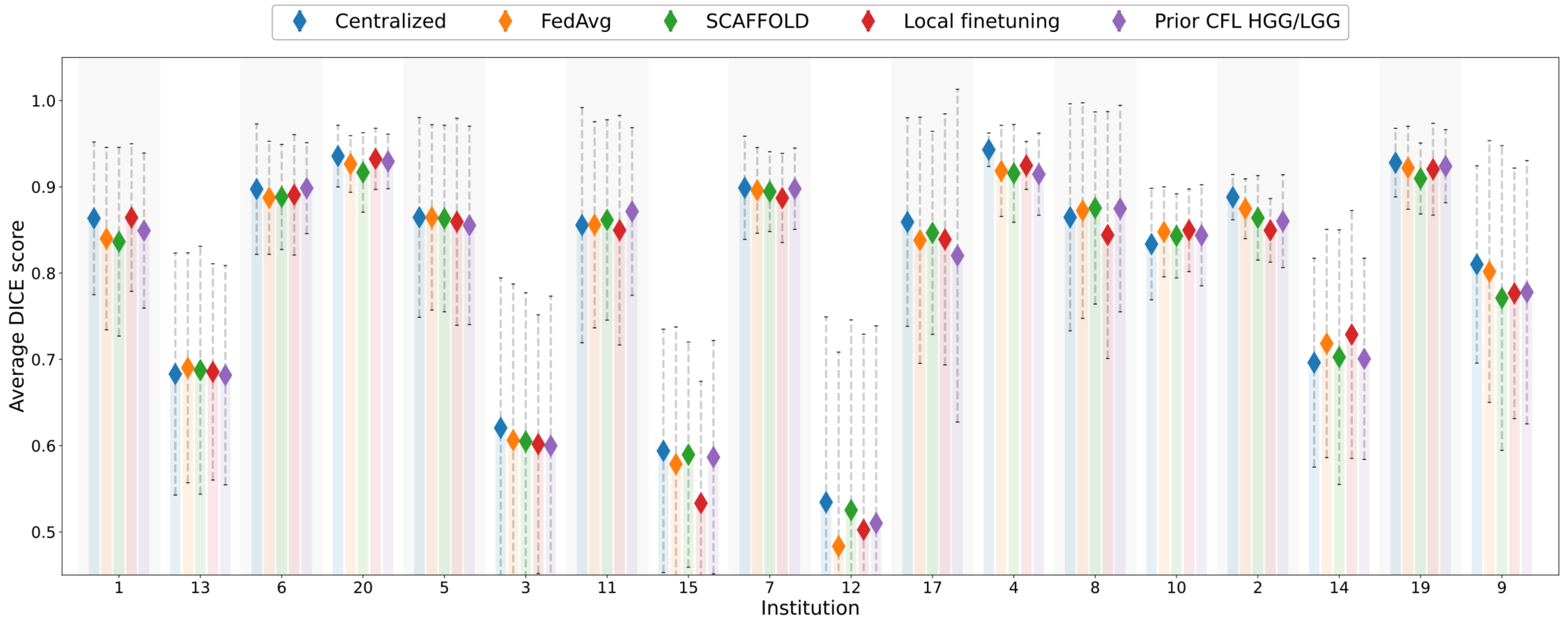}
\caption{Average Dice scores of \textit{centralized} and \textit{FedAvg} training in \textit{Limited data} setup of FeTS2022 dataset per institution, in decreasing order of number of test samples (c.f. Figure \ref{fig:tumor_type_skew_fets2022}). Errors bars represent $\pm$ one standard deviation.}
\label{figknown_tumor_inst}
\end{figure*}
\end{document}